\pdfoutput=1
\documentclass{article}

\usepackage[main, final]{neurips_2026}
\usepackage{tikz}
\usetikzlibrary{positioning,calc,arrows.meta}  %

\usepackage[utf8]{inputenc} %
\usepackage[T1]{fontenc}    %
\usepackage{hyperref}       %
\usepackage{url}            %
\usepackage{booktabs}       %
\usepackage{array}          %
\usepackage{graphicx}       %
\usepackage{float}          %
\usepackage{wrapfig}        %
\usepackage{amsfonts}%
\usepackage{amsmath}
\usepackage{nicefrac}       %
\usepackage{microtype}      %
\usepackage{xcolor}         %
\usepackage{amssymb}
\usepackage{subcaption}     %

\title{Language Models Solve a Classic Theory-of-Mind Task Using Domain-General Mechanisms for Divergence Detection}

\author{%
  Idil Karen Sahin \\
  Dartmouth College, Stanford University \\
  \texttt{idilkarensahin@gmail.com} \\
  \And
  Taylor W. Webb\thanks{Equal advising.} \\
  Princeton University \\
  \And
  Steven M. Frankland\footnotemark[1] \\
  Dartmouth College \\
}

\begin{document}

\maketitle

\begin{abstract}

Large Language Models (LLMs) successfully solve tasks that require reasoning about the truth or falsity of an agent's beliefs, classically used to assess whether a system has theory of mind. However, how LLMs solve these tasks remains unclear. Do they deploy circuitry specialized for reasoning about other minds, memorize common false-belief vignette structures and outputs, or rely on more domain-general computations? We investigate this question primarily in Qwen2.5-14B-Instruct, with selected analyses in Gemma-2-27B-it, using causal mediation and representational similarity analyses across matched prompt sets. A population of mid-layer attention heads tracks divergence between a representation and the current state of the world. A distinct population of later-layer heads retrieves the answer token, and the two populations combine compositionally to determine the output. Notably, this signature holds across structurally similar tasks that differ heavily in content: tasks that require no perspective-taking (true vs. false photographs) and tasks that neither describe situations nor use natural language (aliasing vs. copying variables in python code). These findings show the importance of domain-general contributions to classic false-belief performance in LLMs, while leaving open additional belief-specific computations.

\end{abstract}

\section{Introduction}

Reasoning about the intentions, goals, and beliefs of other people is a foundation of social intelligence. In humans, this capability — known as Theory of Mind (ToM) \citep{premackWoodruff1978} — has classically been assessed using the false-belief (“Sally–Anne”) task \citep{wimmerPerner1983, baronCohen1985}. A character (Sally) forms a belief about the world (her marble is in the basket), loses perceptual access (she leaves the room), and the world changes in her absence (Anne moves the marble to a box). The participant must report what Sally will expect when she returns. The task is constructed so that the correct answer tracks the character’s belief rather than the current state of the world. Success is therefore standardly taken as evidence that a system has ToM (\citep{dennett1978},but see \citep{bloomGerman2000, phillipsKnowledgeBeforeBelief2021}).

Large language models now solve such tasks \citep{kosinskiEvaluatingLargeLanguage2024}, but the mechanisms underlying this success are unclear for two reasons. First, false-belief vignettes are common in training corpora, so success could reflect memorized stories, or a learned template of their event structure, rather than a general ability. Second, performance degrades sharply under small variations of the task \citep{ullman2023, sclarExploreTheoryMind2024}. For example, \citet{sclarExploreTheoryMind2024} generated stories that independently vary characters entering, leaving, communicating, and moving objects; on adversarially sampled stories, accuracy falls to 9\% for GPT-4o and 0\% for Llama-3.1-70B.  Whether this fragility actually distinguishes models from humans remains unsettled, however, as previous work has found that adults also sometimes fail on perturbed variants \citep{strachan2024}. Regardless, what these results do make clear is that behavioral success alone cannot reveal how the task is being solved.

In humans, by contrast, the underlying computation has received considerable study, and decades of cognitive neuroscience suggest it is supported by dedicated neural machinery \citep{saxekanwisher2003, schurz2014}. Neuroimaging consistently implicates two cortical regions — dorsal medial prefrontal cortex (dmPFC) and the right temporoparietal junction (rTPJ) — in mental-state reasoning over closely matched control tasks \citep{frithNeuralBasisMentalizing2006b}. rTPJ, specifically, responds substantially more to false beliefs than to the same content carried by non-mental representations — false photographs, maps, or signs \citep[though see][]{saxe2009theory, callejas2011}. rTPJ activity patterns carry the content of the attributed mental state — which situation the agent represents — while remaining indifferent to whether that representation is true or false \citep{bioRightTemporoparietalJunction2022}. On the standard interpretation, human ToM is supported by a content-sensitive subsystem for reasoning effectively about mental states.

Here, we ask what computations underlie this success in LLMs. These could in principle be specialized for belief tracking, could reduce to retrieval of memorized vignette structure, or could be shared with domain-general state tracking (or some combination thereof). We adjudicate between these possibilities by examining the underlying mechanisms in Qwen2.5-14B-Instruct and Gemma-2-27B-it. Specifically, we use the causal-mediation framework of \citet{yangEmergentSymbolicMechanisms2025} to measure, and intervene on, the contribution of every attention head.

\noindent\begin{minipage}[t]{0.47\textwidth}
\vspace{0pt}

Across matched prompt pairs that vary whether a character’s representation diverges from the world, which answer is correct, or both, we identify two functionally dissociable head populations. A mid-layer population responds to divergence between a representation and the current state of the world, and does so regardless of whether that representation is a character’s belief, a photograph \citep{zaitchik1990}, or a variable in Python code. A late-layer population then retrieves the answer. These two computations compose: when divergence and answer change together, the per-head intervention effects are closely predicted by the unweighted sum of the effects in the two single-change (divergence, answer) conditions ($r = 0.98$ in Qwen; similarly in Gemma). Together, these results suggest that false-belief performance is compositionally assembled from reusable parts, including a domain-general mechanism for divergence detection. 

\end{minipage}\hfill
\begin{minipage}[t]{0.5\textwidth}
\vspace{0pt}

\definecolor{cCaseD}{HTML}{2C6FB5}   %
\definecolor{cCaseN}{HTML}{C8650D}   %
\definecolor{fOneToken}{HTML}{EDEDE8}
\definecolor{fOneAxis}{gray}{0.62}
\definecolor{fOneLine}{gray}{0.15}
\definecolor{fOneFill}{HTML}{F6F6F6}

\captionsetup{type=figure}
\centering
\resizebox{\linewidth}{!}{%
\begin{tikzpicture}[
  base/.style      = {font=\Large, align=center, inner sep=3pt},
  textbox/.style   = {base, draw=none, fill=fOneToken, rounded corners=4pt},
  token/.style     = {textbox, minimum width=3.05cm, minimum height=1.7cm},
  unseen/.style    = {draw=cCaseD, dashed, line width=1.1pt},
  component/.style = {base, draw=fOneLine, fill=fOneFill, rounded corners=4pt,
                      minimum width=3.3cm, inner ysep=5pt, line width=1pt},
  varlabel/.style  = {base, font=\Large\itshape, fill=fOneFill, rounded corners=4pt,
                      inner xsep=8pt, inner ysep=4pt},
  barb/.style      = {-{Straight Barb[length=4pt,width=7pt]}},
  flow/.style      = {barb, fOneLine, line width=1pt},
  route/.style     = {barb, line width=1.4pt, rounded corners=8pt},
  axis/.style      = {-{Straight Barb[length=3.5pt,width=6pt]}, fOneAxis!80, line width=0.5pt},
  axislabel/.style = {font=\Large\itshape, text=fOneAxis},
]
  \def\fOneSep{3.25}    %
  \def\fOneGap{0.5cm}     %
  \def\fOneRouteA{0.2}    %
  \def\fOneRouteB{-0.2}   %

  \node[token] (t1) at (0*\fOneSep,0) {truck\\parked at\\\textbf{Riverside}};
  \node[token] (t2) at (1*\fOneSep,0) {Alice\\\textcolor{cCaseD}{\textbf{looks away}} /\\\textcolor{cCaseN}{\textbf{keeps looking}}};
  \node[token, unseen] (t3) at (2*\fOneSep,0) {truck\\moves to\\\textbf{Sycamore}};
  \node[token] (t4) at (3*\fOneSep,0) {Where will\\Alice look?};

  \node[component, above=\fOneGap of t4] (mid)
    {\textbf{divergence heads}\\\textcolor{gray}{\textit{mid layers}}\\[1pt]
     {\large\textcolor{gray}{belief $\cdot$ photo $\cdot$ code}}};
  \node[varlabel, above=\fOneGap of mid] (state)
    {\textcolor{cCaseD}{\textbf{diverged}} / \textcolor{cCaseN}{\textbf{not diverged}}};
  \node[component, above=\fOneGap of state] (late)
    {\textbf{retrieval heads}\\\textcolor{gray}{\textit{late layers}}};
  \node[textbox, inner xsep=6pt, inner ysep=5pt, above=\fOneGap of late] (out)
    {\textcolor{cCaseD}{\textbf{Riverside}} / \textcolor{cCaseN}{\textbf{Sycamore}}};

  \draw[flow] (t4.north)    -- (mid.south);
  \draw[flow] (mid.north)   -- (state.south);
  \draw[flow] (state.north) -- (late.south);
  \draw[flow] (late.north)  -- (out.south);

  \draw[route, cCaseD] (t1.north) |- ($(late.west)+(0,\fOneRouteA)$);
  \draw[route, cCaseN] (t3.north) |- ($(late.west)+(0,\fOneRouteB)$);

  \coordinate (xA) at ($(t1.south west)+(0,-0.22)$);
  \coordinate (xB) at ($(t4.south east)+(0.05,-0.22)$);
  \coordinate (yA) at ($(t1.north west)+(-0.15,0.1)$);
  \coordinate (yB) at (yA |- out.north);
  \draw[axis] (xA) -- (xB)
    node[axislabel, pos=0, anchor=north west, yshift=-0.08cm] {sequence position};
  \draw[axis] (yA) -- (yB)
    node[axislabel, pos=0, rotate=90, anchor=south west, xshift=-0.08cm] {model depth};
\end{tikzpicture}}
\caption{Mechanism. Alice looks away (blue) or keeps looking (orange); the dashed outline marks the move she misses in the blue case. At the query token, mid-layer divergence heads signal whether her belief has diverged from the world. Later retrieval heads attend to the original or the updated location.}
\label{fig:mechanism}

\end{minipage}\par\medskip
\suppressfloats[t] %
\section{Methods}

\paragraph{Prompt-pair design.} Following the causal mediation framework of \citet{yangEmergentSymbolicMechanisms2025}, all measurements are taken on prompt pairs $(p^{\text{src}}, p^{\text{tgt}})$ constructed to vary divergence status, the correct answer, or both, while matching the remaining task features. The design has nine main conditions, each named by what changes between the two prompts (\textit{divergence-change}, \textit{answer-change}, or \textit{dual-change}) and by its domain (\textit{belief}, \textit{photo}, or \textit{code}), yielding combinations such as \textit{dual-change photo}, \textit{divergence-change code}, and so on, plus a tenth, state-only control (an answer-change condition in a story with no agent, used in Appendix~\ref{sec:appendix-heatmaps-all}). The \textit{divergence-change} condition changes whether the queried representation keeps its original value after the world changes, with the answer held fixed (Figure~\ref{fig:design}b, top); an \textit{answer-change} condition changes the answer with divergence held fixed (middle); a \textit{dual-change} condition changes both (bottom). For the belief and photo tasks, the six story conditions and the state-only control are crossed with three scenario templates (food truck, hair styling, and library book) and two patch directions (diverged~$\to$~not diverged and the reverse direction; in the answer-change conditions, where both prompts share their divergence status, the two directions exchange which location assignment is the source), producing $7 \times 3 \times 2 = 42$ story cells. In code (Figure~\ref{fig:design}a, bottom row), \texttt{b = a.copy()} keeps the original value after \texttt{a[i]} changes and \texttt{b = a} does not, so the three code conditions pair copy with alias (divergence-change; values exchanged so the answers match), copy with copy or alias with alias (answer-change; values exchanged so the answers differ), and copy with alias (dual-change; same values, different answers).
We use three code conditions on a single Python list template, each mirroring a story contrast, which adds $3 \times 2 = 6$ cells (both patch directions for the divergence-change and dual-change conditions; one copy and one alias frame for the answer-change condition), for $48$ cells with $n \in [129, 198]$ valid prompt pairs per cell (per-condition ranges in Table~\ref{tab:appendix-heatmaps}; the \textit{divergence-change belief} and \textit{dual-change belief} cells pool two sampling seeds, and all other cells have at most $150$ pairs). The conditions are illustrated in Figure~\ref{fig:design}; Figure~\ref{fig:stimuli} in Appendix~\ref{sec:appendix-stimuli} gives a full example prompt for every template and domain.

\definecolor{cCaseD}{HTML}{2C6FB5}   %
\definecolor{cCaseN}{HTML}{C8650D}   %
\definecolor{cPaper}{HTML}{F6F6F3}
\providecommand{\figtx}{\fontsize{6.3}{7.3}\selectfont}
\providecommand{\fighd}{\fontsize{6.6}{7.6}\selectfont}
\providecommand{\figbreak}{\tikz[baseline=-0.55ex]{\draw[cCaseD, line width=0.5pt]
  (0,-0.075) -- (0.06,0.075) (0.075,-0.075) -- (0.135,0.075);}}
\providecommand{\figarrD}{\tikz[baseline=-0.55ex]{\draw[->, cCaseD, line width=0.5pt] (0,0) -- (0.46,0);}}
\providecommand{\figarrN}{\tikz[baseline=-0.55ex]{\draw[->, cCaseN, line width=0.5pt] (0,0) -- (0.46,0);}}
\providecommand{\figAnsD}[1]{\textcolor{cCaseD}{\textbf{#1}}}
\providecommand{\figAnsN}[1]{\textcolor{cCaseN}{\textbf{#1}}}
\providecommand{\figAnsP}[1]{\tikz[baseline=(p.base)]\node[draw=black!45, line width=0.35pt,
  rounded corners=2pt, inner sep=1.3pt] (p) {\textbf{#1}};}

\tikzset{
  bx/.style={draw=black!45, line width=0.35pt, rounded corners=2pt,
             inner sep=3pt, align=left},
  arw/.style={->, black!55, line width=0.4pt, shorten >=1pt, shorten <=1pt},
}

\begin{figure}[t]
\centering
\begin{minipage}[c]{0.585\textwidth}
\begin{tikzpicture}[x=1cm, y=1cm]
  \def\xr{0.62}   %
  \def\xw{4.31}   %
  \def\xq{5.88}   %

  \node[font=\small\bfseries, anchor=west, inner sep=0pt] at (0.02,3.12) {(a)};

  \node[font=\fighd\itshape, text=black!55, anchor=west] at (\xr,3.12) {representation formed};
  \node[font=\fighd\itshape, text=black!55, anchor=west] at (\xw,3.12) {world changes};
  \node[font=\fighd\itshape, text=black!55, anchor=west] at (\xq,3.12) {query};

  \foreach \yc/\lab/\rep/\wld/\qry in {%
    2.45/belief/{Alice sees the truck on \textbf{Riverside}, then
                 \textcolor{cCaseD}{\textbf{looks away}} /
                 \textcolor{cCaseN}{\textbf{keeps watching}}}/%
         {the truck moves to \textbf{Sycamore}}/{Where will Alice look?},
    1.32/photo/{a photo of the truck on \textbf{Riverside}, taken
                \textcolor{cCaseD}{\textbf{before}} /
                \textcolor{cCaseN}{\textbf{after}} the move}/%
         {the truck moves to \textbf{Sycamore}}/{Where is the truck in the photo?}}
  {
    \node[font=\figtx\bfseries, anchor=east] at (0.52,\yc) {\lab};
    \node[bx, text width=3.34cm, minimum height=0.86cm, anchor=west,
          font=\figtx] (rep) at (\xr,\yc) {\rep};
    \node[bx, fill=cPaper, text width=1.22cm, minimum height=0.86cm,
          anchor=west, font=\figtx] (wld) at (\xw,\yc) {\wld};
    \node[bx, text width=1.72cm, minimum height=0.86cm, anchor=west,
          font=\figtx] (qry) at (\xq,\yc) {\qry};
    \draw[arw] (rep.east) -- (wld.west);
    \draw[arw] (wld.east) -- (qry.west);
  }

  \node[font=\figtx\bfseries, anchor=east] at (0.52,0.19) {code};
  \node[bx, text width=3.34cm, minimum height=0.86cm, anchor=west,
        font=\figtx] (rep3) at (\xr,0.19)
       {\texttt{a = [..,\textbf{'R'},..]}\\
        \textcolor{cCaseD}{\textbf{\texttt{b = a.copy()}}} /
        \textcolor{cCaseN}{\textbf{\texttt{b = a}}}};
  \node[bx, fill=cPaper, text width=1.22cm, minimum height=0.86cm,
        anchor=west, font=\figtx] (wld3) at (\xw,0.19) {\texttt{a[i] = \textbf{'S'}}};
  \node[bx, text width=1.72cm, minimum height=0.86cm, anchor=west,
        font=\figtx] (qry3) at (\xq,0.19) {\texttt{b[i]}?};
  \draw[arw] (rep3.east) -- (wld3.west);
  \draw[arw] (wld3.east) -- (qry3.west);

  \node[font=\figtx, anchor=west, text width=7.0cm] at (0.62,-0.70)
       {\figarrD\ \,\textcolor{cCaseD}{\textbf{diverged}}: the query returns the original value;\\
        \figarrN\ \,\textcolor{cCaseN}{\textbf{not diverged}}: it returns the updated one.};
\end{tikzpicture}
\end{minipage}\hfill
\begin{minipage}[c]{0.40\textwidth}
\renewcommand{\arraystretch}{1.16}
\figtx
\begin{tabular}{@{}r@{\hspace{5pt}}c@{\hspace{2pt}}c@{\hspace{2pt}}c@{\,$\Rightarrow$\,}c@{}}
\multicolumn{1}{@{}l}{\normalfont\small\bfseries (b)} & \textit{\color{black!55}starts at} & & \textit{\color{black!55}ends at} & \textit{\color{black!55}answer} \\[1pt]
\multicolumn{5}{@{}l}{\fighd\textbf{Divergence change}\enspace{\color{black!55}\itshape answer held}} \\
\textit{source} & Riverside & \figarrD & Sycamore & \figAnsD{Riverside} \\
\textit{target} & Sycamore & \figarrN & Riverside & \figAnsN{Riverside} \\
\textit{target\,+\,patch} & & & & \figAnsP{Sycamore} \\[3pt]
\multicolumn{5}{@{}l}{\fighd\textbf{Answer change}\enspace{\color{black!55}\itshape divergence held}} \\
\textit{source} & Riverside & \figarrD & Sycamore & \figAnsD{Riverside} \\
\textit{target} & Sycamore & \figarrD & Riverside & \figAnsD{Sycamore} \\
\textit{target\,+\,patch} & & & & \figAnsP{Riverside} \\[3pt]
\multicolumn{5}{@{}l}{\fighd\textbf{Dual change}} \\
\textit{source} & Riverside & \figarrD & Sycamore & \figAnsD{Riverside} \\
\textit{target} & Riverside & \figarrN & Sycamore & \figAnsN{Sycamore} \\
\textit{target\,+\,patch} & & & & \figAnsP{Riverside} \\
\end{tabular}
\end{minipage}

\caption{\textbf{(a)} Experiment domains. Each prompt forms a representation, changes the world and queries the representation. \textbf{(b)} The three condition types, shown on the belief example. For each attention head, its output on the source prompt replaces its output on the target prompt; if the head carries the relevant property, this moves the model's answer to the causal answer in the black box.}
\label{fig:design}
\end{figure}
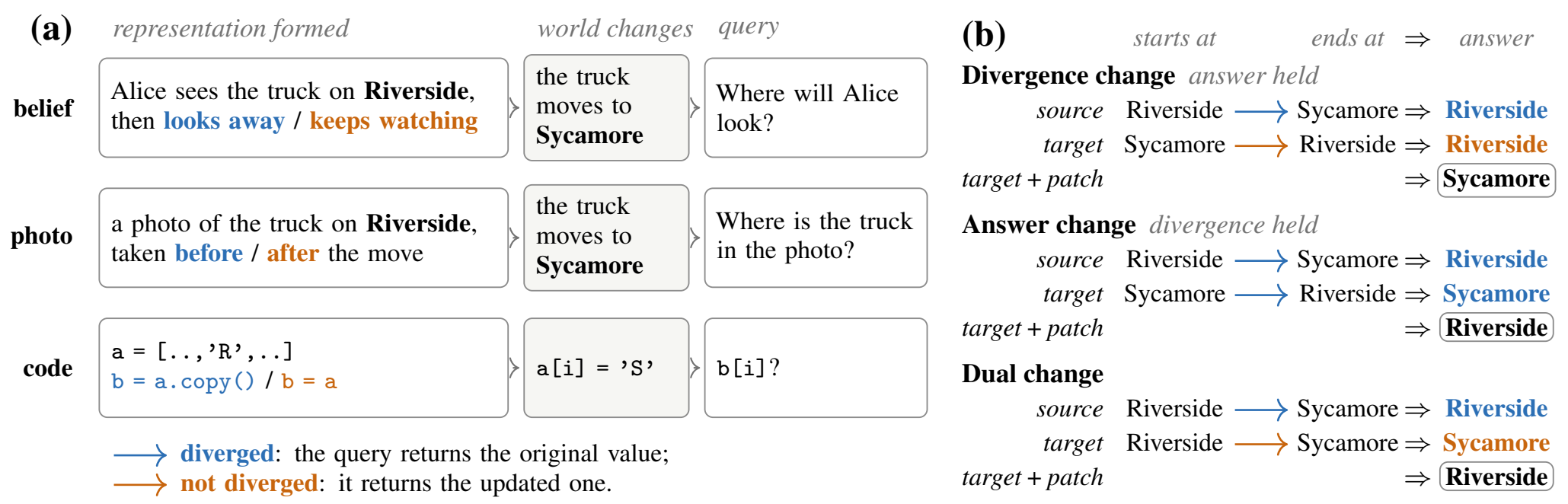

\paragraph{Causal mediation score.} We cache each head's output $z^{(\ell,h)}$ at the final prompt token in $p^{\mathrm{src}}$ and substitute it at the same position in $p^{\mathrm{tgt}}$, yielding $p^{\mathrm{tgt},*}$. Following \citet{yangEmergentSymbolicMechanisms2025} (Eq.~5; Appendix~B.2, Eq.~7), let $y_{\mathrm{tgt}}$ be the target's correct answer and $y_{\mathrm{causal}}$ the answer predicted by the hypothesized intervention. The causal mediation score is
\[
s^{(\ell,h)} =
\bigl(f(p^{\mathrm{tgt},*})[y_{\mathrm{causal}}]-f(p^{\mathrm{tgt},*})[y_{\mathrm{tgt}}]\bigr)
-
\bigl(f(p^{\mathrm{tgt}})[y_{\mathrm{causal}}]-f(p^{\mathrm{tgt}})[y_{\mathrm{tgt}}]\bigr).
\]
Here $f(p)[y]$ is the next-token logit for answer $y$. The score measures how much patching one head changes the logit difference between the predicted intervention answer and the target's correct answer. In divergence-change conditions, the predicted intervention answer (causal answer) is the target's alternative location, not the source's correct answer. Positive scores in all cases indicate a shift toward the causal answer.

\paragraph{Fingerprint comparisons.} We average the intervention scores into one vector per condition, with one entry per attention head; we call this vector a "fingerprint". Pearson correlation between two fingerprints measures whether the same heads have similar patterns of effects across conditions (range $-1$ to $1$). We report signed (raw) scores and, as a sensitivity check, scores with negative cell means set to zero (clamped). Appendix~\ref{sec:appendix-h2} describes the averaging, uncertainty estimates, and permutation tests; Appendix~\ref{sec:steering-methods} describes the interventions.

\paragraph{Hypotheses.} We organize our experiments around the following hypotheses:

\textbf{Compositionality.} Divergence tracking and answer retrieval should be implemented by largely distinct populations of attention heads with different depth distributions, and these two populations should compose together to jointly implement both processes. Concretely, this predicts that 1) the causal fingerprints for the divergence-change and answer-change conditions should reveal largely distinct populations of heads, with 2) the answer-change heads appearing in later layers, and 3) the fingerprints for these conditions should combine additively in the dual-change condition, i.e., the sum of the divergence-change and answer-change fingerprints should be highly correlated with the dual-change fingerprint. 

\begin{figure}[t]
  \centering
  \includegraphics[width=0.9\textwidth]{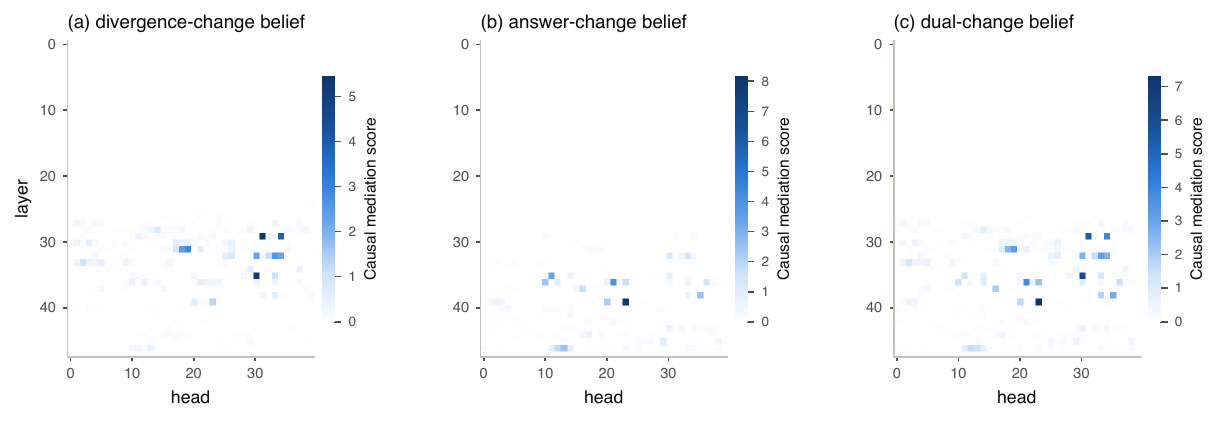}
  \caption{Condition-averaged per-head causal mediation scores in Qwen2.5-14B-Instruct (48 layers $\times$ 40 heads) for \textbf{(a)} \textit{divergence-change belief}, \textbf{(b)} \textit{answer-change belief}, and \textbf{(c)} \textit{dual-change belief}. Negative scores clamped to zero; each panel normalized to its own maximum.}
  \label{fig:cma-belief}
\end{figure}

\textbf{Domain generality.} Divergence tracking should be implemented by the same set of heads across the three task domains (belief, photo, and code). Concretely, this predicts that the \textit{divergence-change belief}, photo, and code fingerprints should be highly correlated. An even stronger test of domain-generality involves patching the outputs of the divergence heads between task domains. For example, patching the output of the divergence heads from a divergence belief trial into a no-divergence photo trial should induce a response consistent with a divergence photo trial.

\textbf{Indirection.} Retrieval heads should retrieve the answer from the appropriate location based on the information computed by the divergence heads. The divergence signal can therefore be thought of as a pointer that refers to the location of the answer rather than the answer itself, thus implementing a form of indirection. Concretely, this predicts that editing the residual stream before the retrieval heads so as to alter the divergence signal should change the attention distribution of those heads. We also investigate the representational structure of the divergence signal, finding that it is best interpreted as an abstract pointer (i.e., coding the abstract role that the answer plays) rather than a precise representation of the answer position.

\section{Results}

\subsection{Distinct populations for divergence detection and answer retrieval}

\noindent\begin{minipage}[t]{0.58\textwidth}
\vspace{0pt}
Do changes in divergence and changes in the answer recruit different attention heads? We compare the per-head causal-mediation fingerprints for \textit{divergence-change belief}, which changes divergence while holding the answer fixed, and \textit{answer-change belief}, which changes the answer while holding divergence fixed. In Qwen2.5-14B-Instruct, the prominent \textit{divergence-change belief} cluster lies at layers 29--35, whereas the prominent \textit{answer-change belief} cluster lies later, at layers 36--46 (Figure~\ref{fig:cma-belief}a,b; see also Figures~\ref{fig:appendix-qwen-c14} and~\ref{fig:appendix-qwen-c57} in Appendix). Their fingerprints are only weakly correlated, with a Pearson correlation of $r=0.260$ with negative scores clamped to zero and $r=0.209$ for raw scores. In Qwen2.5-14B-Instruct, the divergence-change and answer-change head sets shared five heads in the belief task and one head each in the photograph and code tasks. The same qualitative pattern appears in Gemma-2-27B-it: the \textit{divergence-change belief} heads lie in a mid-to-upper band and the \textit{answer-change belief} heads above them (Figure~\ref{fig:appendix-gemma}; $r=0.25$ between the two clamped fingerprints). We next test whether combining the separate divergence-change and answer-change fingerprints predicts the dual-change fingerprints.

\end{minipage}\hfill
\begin{minipage}[t]{0.39\textwidth}
\vspace{0pt}
\captionsetup{type=figure}
  \centering
  \includegraphics[width=\linewidth]{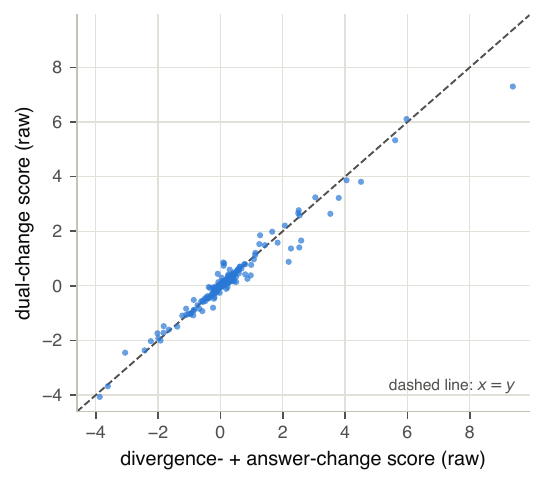}
  \caption{Additive decomposition of the \textit{dual-change belief} condition into \textit{divergence-change belief} and \textit{answer-change belief}. Results reflect correlation between \textit{dual-change belief} fingerprint and sum of divergence-change and answer-change fingerprints for Qwen2.5-14B-Instruct over all 1920 heads. See also: Appendix~\ref{sec:appendix-h2}.}
  \label{fig:c3-decomp}

\end{minipage}\par\medskip

\subsubsection{Compositional combination of divergence detection and answer retrieval}
\label{sec:modularized}

Figure~\ref{fig:cma-belief}c shows the set of heads that are causally involved in the \textit{dual-change belief} condition, in which both divergence and answer change. Visual inspection suggests that this set reflects the combination of the divergence-change and answer-change heads. To confirm this, we performed a correlation analysis. Specifically, if the two populations' intervention effects combine approximately additively, the sum of the \textit{divergence-change belief} and \textit{answer-change belief} fingerprints ($\mathbf{x}_{\mathrm{div}} + \mathbf{x}_{\mathrm{ans}}$) should be highly correlated with the \textit{dual-change belief} fingerprint ($\mathbf{x}_{\mathrm{dual}}$), more so than either component alone:

\begin{equation}
    r(\mathbf{x}_{\mathrm{dual}},\, \mathbf{x}_{\mathrm{div}} + \mathbf{x}_{\mathrm{ans}}) \;>\; \max\bigl(r(\mathbf{x}_{\mathrm{dual}}, \mathbf{x}_{\mathrm{div}}),\, r(\mathbf{x}_{\mathrm{dual}}, \mathbf{x}_{\mathrm{ans}})\bigr).
\end{equation}

For raw fingerprints, the \textit{dual-change belief} fingerprint correlates with the \textit{divergence-change belief} fingerprint at $r=0.807$, the \textit{answer-change belief} fingerprint at $r=0.720$, and their unweighted sum at $r=0.980$. The sum's gain over the better component ($\mathbf{x}_{\mathrm{div}}$) is $0.173$ (95\% head-resampling interval $[0.077,0.220]$, 10,000 resamples). These intervals measure sensitivity to which heads are included, rather than uncertainty over new prompts. When clamping negative scores at zero, the correlations are $0.834$, $0.714$, and $0.971$, and the gain is $0.137$ $[0.023,0.189]$ (Appendix Figure~\ref{fig:c3-specificity}). 

We also observed a similar result in Gemma ($r(\mathbf{x}_{\mathrm{dual}},\, \mathbf{x}_{\mathrm{div}} + \mathbf{x}_{\mathrm{ans}}) =0.986$ for raw scores and $0.985$ for clamped scores over 1472 heads). Together, these results support largely distinct but partially overlapping populations for divergence tracking and answer retrieval, whose causal fingerprints combine approximately additively in the dual-change condition.

\WFclear
\subsection{Evaluating the domain-generality of divergence detection}

\label{sec:cma-fingerprints}

\begin{figure}[H]
  \centering
  \includegraphics[width=\textwidth]{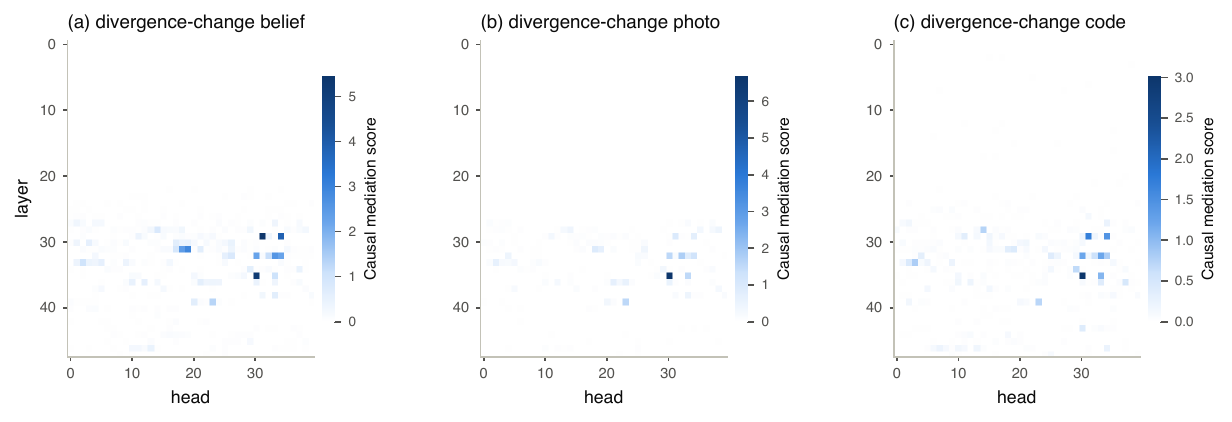}
  \caption{Condition-averaged per-head causal mediation scores in Qwen2.5-14B-Instruct (48 layers $\times$ 40 heads), for \textbf{(a)} \textit{divergence-change belief}, \textbf{(b)} \textit{divergence-change photo}, and \textbf{(c)} \textit{divergence-change code} (code: Python \texttt{copy} vs.\ \texttt{alias}). Negative scores clamped to zero; each panel normalized to its own maximum.}
  \label{fig:cma-overview}
\end{figure}

We next asked: Is divergence detection performed using domain-general mechanisms common to all three tasks? Figure~\ref{fig:cma-overview}a--c shows the divergence-change fingerprints for the belief, photograph, and code tasks, respectively. Visual inspection reveals a significant degree of overlap between these three conditions. Indeed the causal mediation fingerprints are highly correlated across these conditions (Pearson correlation with negative scores clamped at zero: belief--code, $r=0.902$; photograph--code, $r=0.848$; belief--photograph, $r=0.757$; with raw scores: $r=0.867$, $r=0.851$, $r=0.763$, respectively). Similarly high correlations are observed for the answer-change conditions (Pearson correlation for clamped scores: belief--photograph, $r=0.974$; belief--code, $r=0.734$; photograph--code, $r=0.731$; with raw scores: $r=0.957$, $r=0.706$, $r=0.721$, respectively). A state-only control without beliefs or photographs also yielded a causal mediation profile correlated with the answer-change profiles in belief, photograph, and code tasks (Pearson $r=0.907$, $0.949$, and $0.662$ for raw scores, respectively; $0.923$, $0.957$, and $0.681$ with negative scores clamped to zero before averaging). Causal mediation analysis thus identifies very similar sets of heads across tasks.

To test whether the shared fingerprints reflect transferable causal effects, we steered the 22 belief-divergence heads (the top 32 heads of the \textit{divergence-change belief} fingerprint, excluding the top 22 heads of the \textit{answer-change belief} fingerprint; layers 28--36; Appendix~\ref{sec:steering-methods}) using mean activation differences estimated separately from belief, photograph, or code prompts. We added a scaled mean-difference vector to each selected head's output, with $\alpha=+1$ applying an intervention equal in magnitude to that head's mean activation difference in the target task, and $\alpha=-1$ reversing its sign. We tested both positive and negative interventions to determine whether the model's output could be bidirectionally steered, shifting the answer margin toward the alternative answer (for $\alpha=+1$) or toward the original answer (for $\alpha=-1$), in each case measured relative to a norm-matched random direction.

Evaluation uses 72 belief, 72 photograph, and 96 code prompts. For each prompt $p$, we define the full model's output margin after intervention $v$ as
\[
\begin{aligned}
m_v(p)&=f_v(p)[y_{\mathrm{alt}}]-f_v(p)[y_{\mathrm{correct}}],\\
E&=\frac{1}{n}\sum_{p=1}^{n}\bigl[(m_{\mathrm{direction}}(p)-m_{\mathrm{clean}}(p))-(m_{\mathrm{random}}(p)-m_{\mathrm{clean}}(p))\bigr].
\end{aligned}
\]
Here $f_v(p)[y]$ is the final next-token logit for answer $y$, after the remaining layers have processed the edited head outputs. Positive $E$ suggests that the state direction shifts the model toward the alternative answer more than a norm-matched random direction does,  while negative $E$ indicates a shift toward the original answer relative to the norm-matched random direction (Appendix~\ref{sec:steering-methods}).

\begin{figure}[H]
  \centering
  \includegraphics[width=\textwidth]{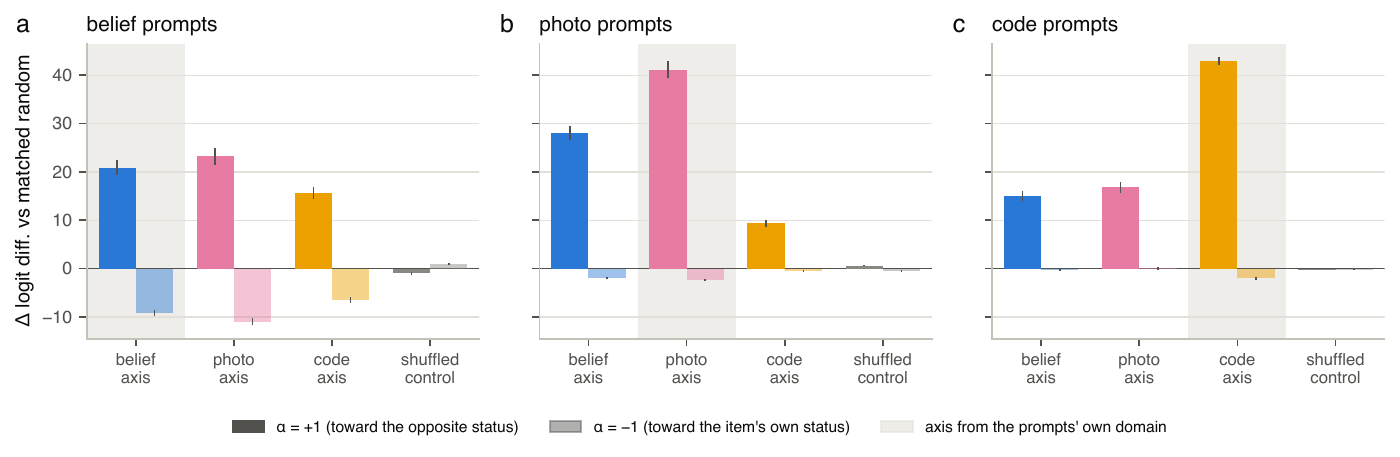}
  \caption{Per-head steering of the 22 belief-divergence heads along mean-difference directions from each task. Change in logit difference vs.\ a matched-norm random direction, sampled independently for each prompt and head; error bars show the standard error of the mean over prompts. Shading marks the direction estimated from the prompts' own task.}
  \label{fig:head-steering}
\end{figure}

Figure~\ref{fig:head-steering} shows the results of these interventions. On belief prompts, photograph-derived directions shift this margin by $+23.2$ logits, compared with $+20.9$ for belief-derived directions and $+15.6$ for code-derived directions ($\alpha=+1$, $n=72$). Reversing the sign ($\alpha=-1$) reverses the effect for all three sources. Transfer also occurs in the other target tasks, but its magnitude varies: on photograph prompts, belief and code directions produce 68\% and 23\% of the photograph-derived effect; on code prompts, belief and photograph directions produce 36\% and 39\% of the code-derived effect. Permuting the same directions across heads, with norms unchanged (`shuffled control' in Figure~\ref{fig:head-steering}), yields effects within $\pm1$ logit. 
These results indicate that the divergence heads encode a shared `divergence' direction that can be effectively transferred across tasks, supporting the \textit{domain-generality} hypothesis.

\subsection{Divergence-related signals drive retrieval via indirection}

Finally, we asked whether divergence-related signals act as a pointer, driving answer retrieval by influencing the position to which retrieval heads attend. To do so, we computed the average difference between false belief trials and true belief trials in the residual stream at the final prompt token after layer 41, the layer immediately prior to the 26 answer retrieval heads (layers 42--46). We then either added or subtracted this direction to the residual stream at the input to layer 42 and measured the attention distribution in the answer retrieval heads. We selected residual-steering coefficient $\beta=4$ from a parameter sweep to balance answer redirection with output stability. Appendix~\ref{sec:steering-methods} reports the sweep, matched intervention controls, and an exploratory output-stability control using 225 belief-free prompts.

\begin{figure}[t]
  \centering
  \includegraphics[width=0.8\textwidth]{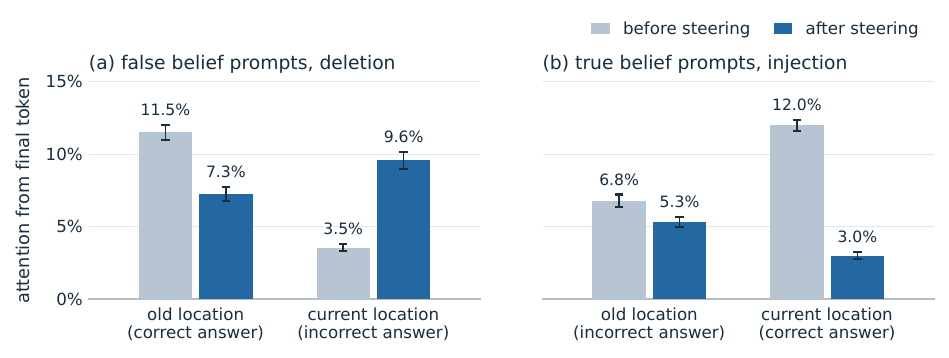}
  \caption{Attention from the final token to the old and current locations, averaged over 26 answer retrieval heads in layers 42--46 of Qwen2.5-14B-Instruct, before and after steering a single residual direction at layer 41 ($\beta = 4$). \textbf{(a)} Removed from false-belief prompts; \textbf{(b)} added to true-belief prompts. Location mass is the sum of attention over token positions overlapping matches to each location's name; attention to all other tokens is omitted, not renormalized (mask sensitivity in Appendix~\ref{sec:steering-methods}). Error bars: 95\% bootstrap intervals over $225$ held-out prompts per condition.}
  \label{fig:steer-attn}
\end{figure}

Figure~\ref{fig:steer-attn} shows the results of this intervention. On false-belief trials (Figure~\ref{fig:steer-attn}a), before steering, the retrieval heads attended primarily to the old location, as expected. After subtracting the false-minus-true direction, attention was directed more to the current location than the old location. The opposite pattern was observed on true belief trials (Figure~\ref{fig:steer-attn}b). Before steering, attention was primarily directed to the current location. After adding the false-minus-true direction, attention was directed more to the old location than the current location.  

To test whether this redirection of attention contributes causally to the model's answer, we intervened directly on the retrieval heads' queries. Resetting these queries moved the model's answer preference back toward its original preference. Transferring the modified queries to an otherwise unaltered model shifted its answer preference in the same direction as altering the divergence signal, but by a smaller amount. The reversal exceeded that of matched control interventions (\hyperref[sec:query-restoration]{Appendix~\ref*{sec:steering-methods}}). These results support the \textit{indirection} hypothesis: the divergence signal influences answer selection in part by directing where the retrieval heads look for information.

\begin{figure}[t]
  \centering
  \includegraphics[width=0.85\textwidth]{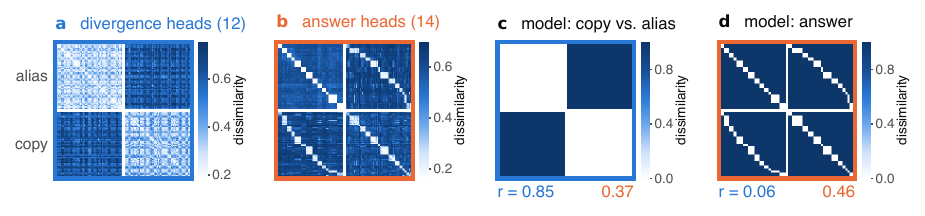}
  \caption{RDMs of the code divergence heads (a) and code answer heads (b) over 144 Python programs, beside the model RDM each set is predicted to fit: copy vs.\ alias (c) and answer identity (d). All panels share one stimulus order and colormap; observed RDMs show cosine distance, model RDMs 0 (same) to 1 (different). Under each model: Mantel correlation with (a) in blue and (b) in orange. Position models in Figure~\ref{fig:rdm-exp1}.}
  \label{fig:rdm-main}
\end{figure}

\paragraph{Abstract divergence versus answer position.} To better understand the nature of the divergence signal and the way in which it might operate as a pointer, we investigated the representational structure of the divergence head outputs. Specifically, we asked whether these heads represent divergence as an abstract variable or whether they represent the precise sequence position of the answer. In our belief and photograph stories, an outdated representation points to the first-mentioned location, so divergence is confounded with answer position. Code stimuli let us vary the queried slot and surrounding text independently of copy/alias (i.e., divergence) status (Appendix~\ref{sec:per-cond-rdm}). We performed representational similarity analysis~\citep{kriegeskorte2008representational}, constructing representational dissimilarity matrices (RDMs) corresponding to idealized representational formats, and comparing these to RDMs constructed from the outputs of the divergence heads and answer retrieval heads. We found that the divergence head RDM (Figure~\ref{fig:rdm-main}a) was a very close match to an RDM for abstract divergence (i.e., simply representing whether a context involves divergence or not; Figure~\ref{fig:rdm-main}c; Spearman $r=0.85$). By contrast, the answer retrieval head RDM (Figure~\ref{fig:rdm-main}b) was a very close match to an RDM for answer identity (Figure~\ref{fig:rdm-main}d; Spearman $r=0.46$). We also compared the divergence head RDM to several RDMs constructed from representations of the sequence position for the answer, and found that these were a much weaker match (Appendix~\ref{sec:per-cond-rdm}). These results suggest that the divergence heads are best thought of as representing an abstract pointer rather than the sequence position of the answer, but further study is required to determine the precise nature of the divergence signal and the way in which it drives the downstream retrieval process.

\section{Discussion}
We asked how LLMs solve classic false-belief tasks, considering three candidate
accounts: retrieval of memorized vignette structure, a specialized
belief-tracking mechanism analogous to the dedicated machinery implicated in
human ToM \citep{saxekanwisher2003, saxe2009theory, schurz2014}, and domain-general state tracking. Our results favor the third. The first -- a
memorization account --predicts machinery tied to the narrative conventions of
false-belief stories. Instead, we find that the head populations that causally support
belief answers are recruited by tasks that involve reasoning about photographs, and even more strikingly, about updated variables in python code. Whereas human findings have been interpreted as evidence for specialized mental-state reasoning \citep{saxekanwisher2003}, our analyses identify a general divergence-tracking computation without identifying a distinct belief-specific mechanism. Steering directions estimated in those domains shift belief answers at
$75$--$111\%$ of the belief-derived effect.  Instead, the evidence here supports a compositional, domain-general solution: a mid-layer computation
detects divergence, a late-layer computation retrieves the answer, and the two
combine approximately additively. These results do not exclude the alternatives
outright --- the regular structure of our stimuli leaves 
template strategies possible (although these templates would have to be quite abstract), and the absence of belief-selectivity in the
mechanisms we identify does not show that \textit{no} circuits are belief-specific. But on these tasks, the computation we were able to identify involves assembling reusable parts, using domain-general mechanisms.

We articulated three hypotheses for what such a system should look like. Focusing on these, the evidence is strongest for
\textit{compositionality} and \textit{domain generality}, and suggestive for
\textit{indirection}. Regarding compositionality, the sum of the divergence-change and answer-change
fingerprints predicts the dual-change fingerprint better than either component
alone, in both models
(\S\ref{sec:modularized}). Regarding domain-generality, divergence fingerprints are highly correlated across
belief, photograph, and code, and directions estimated in each domain steer the
others (Figure~\ref{fig:head-steering}). Regarding indirection, analyses support the central
prediction: altering the divergence signal in the residual stream reorients the retrieval heads' attention between the candidate locations. We note, however, that with only two candidate referents, a classic pointer is hard to distinguish from a bias on the retrieval heads’ queries that has been imposed by the divergence heads. Distinguishing subtly different mechanisms requires a larger address space — three or more plausible candidates, or agent-indexed reference (Alice’s versus Bob’s stored value), which is an important target for future work. Nonetheless, it is reasonable to consider these \textit{abstract} pointers that code for the functional role that the answer plays.

We note that these results connect closely to the lookback mechanism of \citet{prakashLanguageModelsUse2025}. In a lookback, the model copies the same identifier to two places: an \emph{address}, alongside the stored content, and a \emph{pointer}, at the later token where that content will be needed. Retrieval heads then match pointer to address through their query--key circuit, retrieving the stored content. In their belief tasks, a further signal — derived from whether one character can observe another’s actions — controls whether an observed change enters the queried belief. Our divergence signal is also answer-independent and operates on the query side of retrieval as expected for a pointer rather than a retrieved value. Our results add two things to those of Prakash et al. First, the selector is domain-general: it responds to divergence between a representation and the world in photographs and program state, supporting an account in which belief tracking recruits general-purpose addressing machinery. Second, the pointer is abstract rather than precise: our representational analyses indicate the signal encodes which role to consult (the stored versus the current value; object-level divergence rather than the queried slot; Appendix~\ref{sec:per-cond-rdm}). Retrieval heads' own query–key matching resolves the particular token position. A natural hypothesis is that the two are components of a shared higher-order process: the abstract divergence signal determines whether the retrieval heads dereference the stored or the current value, and the retrieval heads’ query–key match, a lookback in the sense of \citet{prakashLanguageModelsUse2025}, carries out that dereferencing of particular tokens.

Our results also align with recent work on linear truth encodings in language models. \citet{ravfogelEmergenceLinearTruth2025} argue that small LMs implicitly acquire a linear axis separating true from false statements because such an axis lowers next-token loss when factual statements co-occur with other factual statements. Our interventions concern a different distinction: whether representations in different domains (belief, photograph, or copied program state) diverge from their updated counterparts. Cross-frame transfer shows that this distinction has a shared causal component, but does not establish that it is the same axis as the truth/falsity of a statement.

\subsection{Limitations and future work}

We note several important limitations to our work.  First, our prompt set covers first-order false-belief tasks instantiated in three scenario templates, but theory-of-mind is a much richer space of constructs. Our results do not address whether the same head population mediates intention attribution, desire inference, or second-order belief reasoning. However, the present results do suggest that the same patterns should hold in other ToM tasks only when success on those tasks involves tracking the divergence between two states of affairs. The pattern appears to follow the domain-general computation, not the content of the task. 

Nonetheless, tracking divergence is relevant for other instances of social reasoning as well.
For example, interactive multiagent studies make the distinction between world state and another agent's information practically relevant. In Decrypto, \citet{lupuDecrypto2025} find that evaluated LLM agents lag behind humans and simple word-embedding baselines in a communication game requiring agents to consider what teammates and opponents know. 
These tests would assess whether the mechanisms found here contribute to emergent multiagent behavior.

We have also not tested repeated belief updates as a character receives new observations or messages. For example, a character might see an object moved to a new location and later be told, incorrectly, that it remains in the old one. Tracking these successive changes goes beyond the single-update structure we studied here. Similarly, nested belief reasoning requires distinguishing whose belief is represented at each level, as in, for example, ``Alice believes that Bob thinks the object is in the drawer.'' \citet{prakashLanguageModelsUse2025} identify character--object--state bindings in first-order tasks,  and \citet{street2025llms} report adult-level performance by GPT-4 on their higher-order ToM test suite. Future mechanistic extensions of our study could test whether the same divergence-sensitive heads operate at different levels of nesting, including how each signal is bound to the relevant agent.

Second, the brittleness exposed by \citet{sclarExploreTheoryMind2024} on procedurally generated ToM stories suggests caution in generalizing our results from the novel, but simple, variants of the classic false-belief structure we use here to more complex, naturalistic ones. Although the generality across tasks we identify is notable, our belief, photo, and code tasks all have a simple structure. We have not yet tested whether intervening on these same heads changes belief answers when stories contain the more varied event sequences generated by ExploreToM. Given the regular structure of our tasks, strategies that need not generalize to flexible reasoning about other minds remain possible. Our results identify a shared mechanism used on these tasks, but do not establish whether the models also perform mentalizing.

Finally, our analyses are restricted to two instruction-tuned models. While the models come from different families and sizes, systematic cross-model and cross-scale verification remains necessary for showing generalization at a broader scale. Whether the divergence-tracking heads we identify emerge during pretraining or are sharpened by post-training is also not addressed by our design. Still, in these models and on these tasks, the relevant mechanisms are neither rote nor richly adapted to the complexity of reasoning about other minds. Instead, they constitute a compositional system of domain-general parts, reused when a representation is checked against the current state of the world.

\paragraph{Code and data availability.} Our code and data for every figure and table are available at \url{https://github.com/idilks/theory-of-mind-cma}.

\begin{ack}
We thank David Kraemer and Soroush Vosoughi for helpful discussions. IKS was supported by the Dartmouth Hanlon Scholars Program. SMF was supported by National Science Foundation (NSF) under Cooperative Agreement No. 2433429, “NSF
AI Research Institute on Interaction for AI Assistants (ARIA)”. 
The authors declare no competing interests.
\end{ack}

\bibliographystyle{unsrtnat}
\bibliography{tom-interp}

\appendix

\section{Example prompts}
\label{sec:appendix-stimuli}

Figure~\ref{fig:stimuli} shows one example of every prompt type: the three story templates in their belief, photograph and state-only versions, and the code template.

\definecolor{cCaseD}{HTML}{2C6FB5}   %
\definecolor{cCaseN}{HTML}{C8650D}   %
\definecolor{cPaper}{HTML}{F6F6F3}
\definecolor{cNone}{gray}{0.62}      %
\providecommand{\figtx}{\fontsize{6.3}{7.3}\selectfont}
\providecommand{\fighd}{\fontsize{6.6}{7.6}\selectfont}
\providecommand{\figAnsD}[1]{\textcolor{cCaseD}{\textbf{#1}}}
\providecommand{\figAnsN}[1]{\textcolor{cCaseN}{\textbf{#1}}}

\newcommand{\stD}[1]{\textcolor{cCaseD}{\textbf{#1}}}
\newcommand{\stN}[1]{\textcolor{cCaseN}{\textbf{#1}}}
\newcommand{\val}[1]{\textbf{#1}}
\newcommand{\stmute}[1]{\textcolor{black!55}{\itshape #1}}
\newcommand{\stvar}[2]{%
  \begin{tabular}[t]{@{\color{#1}\vrule width 1.1pt\hspace{3.5pt}}
                     >{\raggedright\arraybackslash}p{\dimexpr\linewidth-4.6pt\relax}@{}}
  #2
  \end{tabular}}
\newcommand{\stgap}{\par\vspace{3.5pt}}
\newcommand{\stQ}[1]{\rule{0pt}{9.5pt}\stmute{Q:}~#1}
\newcommand{\stA}[1]{\par\vspace{1.5pt}$\Rightarrow$\,#1}
\newcommand{\stkey}[1]{\textcolor{#1}{\rule[-0.5pt]{1.1pt}{6.8pt}}\,}
\newcommand{\dhead}[2]{{\fontsize{8}{9.5}\selectfont\bfseries #1}\newline{\fontsize{6.3}{7.5}\selectfont\stmute{#2}}}
\newcommand{\tplabel}[1]{\multicolumn{3}{@{}p{\textwidth}@{}}{%
  {\fighd\color{black!62}\textsc{#1}}\enspace
  {\color{black!18}\leaders\vrule height 2.6pt depth -2.3pt\hfill}\parfillskip=0pt\par}\\[2pt]}
\newcommand{\tpgap}{\\\noalign{\vspace{7pt}}}
\newcommand{\cq}[1]{\textquotesingle#1\textquotesingle}
\newcommand{\cbox}[1]{%
  \noindent\colorbox{cPaper}{\parbox{\dimexpr\linewidth-2\fboxsep\relax}{\ttfamily#1}}\par\vspace{2pt}}

\begin{figure}[p]
\centering
\figtx
\setlength{\tabcolsep}{0pt}
\renewcommand{\arraystretch}{1}
\newcolumntype{B}{>{\raggedright\arraybackslash}p{0.432\textwidth}}
\newcolumntype{P}{>{\raggedright\arraybackslash}p{0.318\textwidth}}
\newcolumntype{S}{>{\raggedright\arraybackslash}p{0.19\textwidth}}
\newcommand{\cgap}{\hspace{0.03\textwidth}}

\begin{tabular}{@{}B@{\cgap}P@{\cgap}S@{}}
\multicolumn{3}{@{}c@{}}{%
  \stkey{cCaseD}\textcolor{cCaseD}{\textbf{diverged}}: the representation is outdated and same as its original value\qquad
  \stkey{cCaseN}\textcolor{cCaseN}{\textbf{not diverged}}: the representation gets updated\qquad
  \stkey{cNone}\textbf{state only}: no representation}\\[5pt]
\specialrule{0.8pt}{0pt}{3pt}
\dhead{Belief}{an agent's belief vs.\ the world}
  & \dhead{Photo}{a photograph vs.\ the world}
  & \dhead{State only}{the world alone} \\[1pt]
\specialrule{0.4pt}{1pt}{5pt}
\tplabel{Food truck}
\stvar{cCaseD}{Alice sees a food truck selling tacos, on \val{Elm Avenue}. She drives past the food truck and continues down the road and is unable to see the truck. She makes a note to herself to grab a plate of tacos later from the food truck. \stD{After Alice is out of sight,} Bob, the owner of the truck, drives the truck to \val{Juniper Lane}. After this, the food truck stays in place, nobody moves it.}
\stgap
\stvar{cCaseN}{Alice sees a food truck selling tacos, on \val{Elm Avenue}. Bob, the owner of the food truck, drives the truck to \val{Juniper Lane}. \stN{Alice sees Bob moving the truck.} She makes a note to herself to grab a plate of tacos later from the food truck. She drives past the food truck and continues down the road and is unable to see the truck. After this, the food truck stays in place, nobody moves it.}
&
\stvar{cCaseD}{A food truck selling tacos is on \val{Elm Avenue}. A photographer takes a photo of the food truck. \stD{After the photograph is taken,} Bob, the owner, drives the truck to \val{Juniper Lane}. The food truck stays at \val{Juniper Lane}.}
\stgap
\stvar{cCaseN}{A food truck selling tacos is on \val{Elm Avenue}. Bob, the owner, drives the truck to \val{Juniper Lane}. \stN{After the truck is moved, a photographer takes a photo} of the food truck. The food truck stays at \val{Juniper Lane}.}
&
\stvar{cNone}{A food truck selling tacos is on \val{Elm Avenue}. Bob drives the truck to \val{Juniper Lane}. The food truck stays in place. Where is the food truck?}
\\
\stQ{Where would Alice go to find the truck? Answer with only the location, be concise}
  \stA{\figAnsD{Elm Avenue}\,/\,\figAnsN{Juniper Lane}}
&
\stQ{Where does the photograph show the food truck? Answer with only the location, be concise}
  \stA{\figAnsD{Elm Avenue}\,/\,\figAnsN{Juniper Lane}}
&
\stQ{Answer only with the location, be concise.}
  \stA{\textbf{Juniper Lane}}
\tpgap
\tplabel{Hair styling}
\stvar{cCaseD}{Alice and Jennifer are good friends. Alice has \val{brunette} hair and she enjoys dying her hair time and again. On Monday, Alice dyes her hair \val{violet}, but the process goes terribly and she \stD{avoids telling this to Jennifer} and avoids all her friends and does not go outside. Tuesday is Alice's birthday and Jennifer is out shopping for a scarf as a birthday gift for Alice. As hair color affects how accessories look, the sales assistant ask what color Alice's hair is.}
\stgap
\stvar{cCaseN}{Alice and Jennifer are good friends. Alice has \val{brunette} hair and she enjoys dying her hair time and again. On Monday, Alice dyes her hair \val{violet}, but the process goes terribly and she \stN{immediately calls Jennifer to tell her} about the new hair color \val{violet}. Tuesday is Alice's birthday and Jennifer is out shopping for a scarf as a birthday gift for Alice. As hair color affects how accessories look, the sales assistant asks what color Alice's hair is.}
&
\stvar{cCaseD}{Alice has \val{brunette} hair. A photographer takes a photo of Alice for her ID card. \stD{After the photo is taken,} Alice dyes her hair \val{violet}. The hair dye change is permanent and Alice keeps this new color.}
\stgap
\stvar{cCaseN}{Alice has \val{brunette} hair. She dyes her hair \val{violet}. \stN{After her hair is dyed} \val{violet}\stN{, a photographer takes a photo} of Alice for her ID card. The hair dye change is permanent and Alice keeps this new color.}
&
\stvar{cNone}{Alice has \val{brunette} hair. On Monday, Alice dyes her hair \val{violet}. What color is Alice's hair?}
\\
\stQ{What color would Jennifer say Alice's hair is? Answer with only the color, be concise.}
  \stA{\figAnsD{brunette}\,/\,\figAnsN{violet}}
&
\stQ{What hair color does the ID card photo show? Answer with only the color, be concise}
  \stA{\figAnsD{brunette}\,/\,\figAnsN{violet}}
&
\stQ{Answer only with the color, be concise.}
  \stA{\textbf{violet}}
\tpgap
\tplabel{Library book}
\stvar{cCaseD}{Sarah checks out a history book from the \val{fiction} section of the library. She sits down and reads the book in the desk area in \val{fiction}. Later, she finishes the book and leaves to get coffee from the café downstairs before continuing her work. \stD{While she's away,} a librarian sees the book and, thinking it was left behind accidentally, returns it to the \val{reference} section where it actually belongs. The librarian reshelves it properly and it stays there.}
\stgap
\stvar{cCaseN}{Sarah checks out a history book from the \val{fiction} section of the library. She sits down and reads the book in the desk area in \val{fiction}. Sarah sees a librarian approaching and explains she's just getting coffee. The librarian mentions the book should be in \val{reference} section and offers to move it there. Sarah agrees and \stN{watches the librarian take the book} to \val{reference} section. Sarah then leaves to get coffee.}
&
\stvar{cCaseD}{A history book is on a desk in \val{fiction} section of the library. A college photographer takes a picture of the desk area. \stD{After the photograph is taken,} a librarian moves the book to \val{reference} section where it belongs. The book stays in \val{reference} section.}
\stgap
\stvar{cCaseN}{A history book is on a desk in \val{fiction} section of the library. A librarian moves the book to \val{reference} section where it belongs. \stN{After the book is moved, a college photographer takes a photo} of \val{reference}. The book stays in \val{reference} section.}
&
\stvar{cNone}{Sarah places a history book in the \val{fiction} section of the library. A librarian moves the book to the \val{reference} section. The book stays there. Where is the book?}
\\
\stQ{Which section would Sarah look for her book in? Answer with only the section name (eg. young adult, classics), be concise.}
  \stA{\figAnsD{fiction}\,/\,\figAnsN{reference}}
&
\stQ{Which section does the photograph show the book in? Answer with only the section name (eg. young adult or classics), be concise}
  \stA{\figAnsD{fiction}\,/\,\figAnsN{reference}}
&
\stQ{Answer with only the section name (eg. young adult, classics), be concise.}
  \stA{\textbf{reference}}
\\
\specialrule{0.8pt}{8pt}{3pt}
\multicolumn{3}{@{}p{\textwidth}@{}}{\dhead{Code}{an indexed list; a copy diverges from the original list, while an alias does not}}\\[1pt]
\specialrule{0.4pt}{1pt}{5pt}
\multicolumn{3}{@{}p{\textwidth}@{}}{%
\begin{minipage}[t]{0.485\textwidth}
\stvar{cCaseD}{\cbox{a = [\cq{D}, \cq{R}, \cq{8}, \cq{G}, \cq{X}, \cq{D}]\\
        \textcolor{cCaseD}{\textbf{b = a.copy()}}\\
        a[1] = \cq{S}}\vspace{-3pt}
  \stQ{What is b[1]?}\stA{\textcolor{cCaseD}{\ttfamily\bfseries\cq{R}}}}
\end{minipage}\hfill
\begin{minipage}[t]{0.485\textwidth}
\stvar{cCaseN}{\cbox{a = [\cq{D}, \cq{R}, \cq{8}, \cq{G}, \cq{X}, \cq{D}]\\
        \textcolor{cCaseN}{\textbf{b = a}}\\
        a[1] = \cq{S}}\vspace{-3pt}
  \stQ{What is b[1]?}\stA{\textcolor{cCaseN}{\ttfamily\bfseries\cq{S}}}}
\end{minipage}}\\
\specialrule{0.8pt}{4pt}{0pt}
\end{tabular}

\caption{Example prompts for every template and domain with one draw of the variables; the two candidate answers are bold. Colored bars identify divergence status. Q is the question that follows the story; $\Rightarrow$ gives the correct answers. Conditions pair these prompts as in Figure~\ref{fig:design}b.}
\label{fig:stimuli}
\end{figure}

\section{Causal mediation heatmaps across story templates}
\label{sec:appendix-heatmaps-all}

We report causal mediation fingerprints for all seven Qwen story conditions and the four replicated Gemma conditions, with interventions at the final prompt token. Each map averages six cell means (three templates $\times$ two directions), setting negative cell means to zero before averaging. Each color scale runs from zero to that condition's maximum score, $v_{\max}$, so colors cannot be compared directly across conditions. Significance maps mark heads exceeding the family-wise-error-rate (FWER) corrected threshold ($\alpha=0.05$, $5{,}000$ permutations) for the mean score across cells. Table~\ref{tab:appendix-heatmaps} lists the prompt pairs per cell, color-scale maximum, FWER threshold and number of significant heads for each map.

\begin{table}[H]
\centering
\small
\setlength{\tabcolsep}{4pt}
\begin{tabular}{llcccc}
\toprule
Model & Condition & Pairs per cell & $v_{\max}$ & FWER threshold & Significant heads \\
\midrule
Qwen2.5-14B & \textit{divergence-change belief} & 146--198 & 5.44 & 0.42 & 49 of 1920 \\
 & \textit{divergence-change photo} & 139--150 & 6.68 & 0.59 & 19 of 1920 \\
 & \textit{dual-change belief} & 142--198 & 7.30 & 0.47 & 70 of 1920 \\
 & \textit{dual-change photo} & 139--150 & 15.63 & 1.06 & 34 of 1920 \\
 & \textit{answer-change belief} & 142--150 & 8.16 & 0.57 & 34 of 1920 \\
 & \textit{answer-change photo} & 150 & 11.65 & 0.77 & 23 of 1920 \\
 & \textit{answer-change state-only} & 129--150 & 10.58 & 0.77 & 29 of 1920 \\
\midrule
Gemma-2-27B & \textit{divergence-change belief} & 49--198 & 1.24 & 0.16 & 38 of 1472 \\
 & \textit{divergence-change photo} & 150--200 & 1.88 & 0.14 & 47 of 1472 \\
 & \textit{dual-change belief} & 146--150 & 1.51 & 0.15 & 96 of 1472 \\
 & \textit{answer-change belief} & 142--150 & 1.12 & 0.10 & 80 of 1472 \\
\bottomrule
\end{tabular}
\caption{Values for the cross-template averaged maps in Figures~\ref{fig:appendix-qwen-c14}--\ref{fig:appendix-gemma}. Each map averages 6 cells (3 templates $\times$ 2 directions).}
\label{tab:appendix-heatmaps}
\end{table}

\subsection{Qwen2.5-14B-Instruct story conditions}
\label{sec:appendix-heatmaps-qwen}

Figures~\ref{fig:appendix-qwen-c14} and~\ref{fig:appendix-qwen-c57} show the seven Qwen story conditions: the divergence-change and dual-change conditions in Figure~\ref{fig:appendix-qwen-c14}, and the answer-change conditions, including the state-only control, in Figure~\ref{fig:appendix-qwen-c57}.

\begingroup
\renewcommand{\arraystretch}{0.4}
\setlength{\tabcolsep}{2pt}

\begin{figure}[p]
\centering
\begin{subfigure}[t]{0.49\textwidth}
  \includegraphics[width=\linewidth]{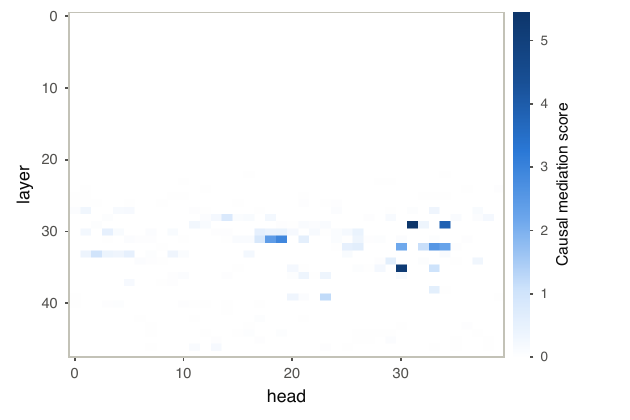}
  \caption{\textit{divergence-change belief}, average score}
\end{subfigure}\hfill
\begin{subfigure}[t]{0.49\textwidth}
  \includegraphics[width=\linewidth]{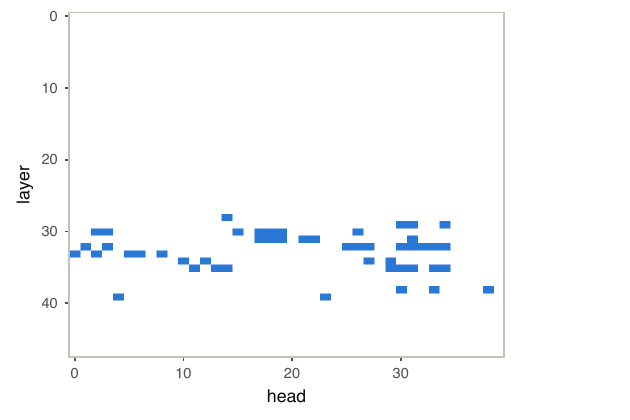}
  \caption{\textit{divergence-change belief}, significant heads}
\end{subfigure}
\par\smallskip
\begin{subfigure}[t]{0.49\textwidth}
  \includegraphics[width=\linewidth]{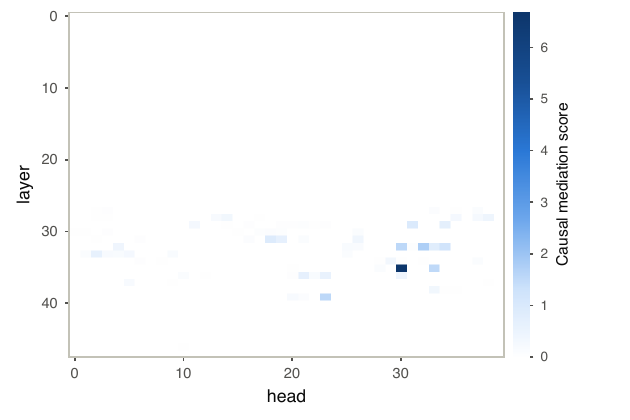}
  \caption{\textit{divergence-change photo}, average score}
\end{subfigure}\hfill
\begin{subfigure}[t]{0.49\textwidth}
  \includegraphics[width=\linewidth]{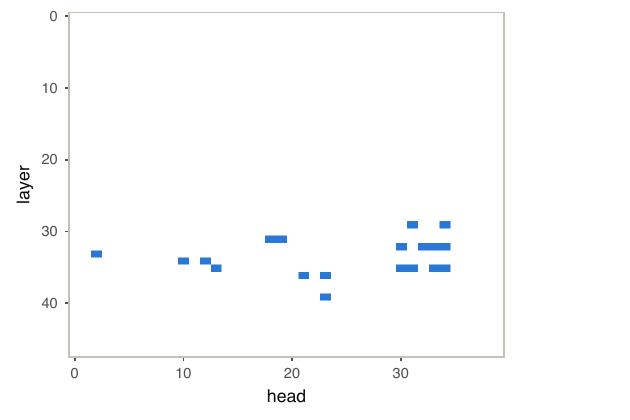}
  \caption{\textit{divergence-change photo}, significant heads}
\end{subfigure}
\par\smallskip
\begin{subfigure}[t]{0.49\textwidth}
  \includegraphics[width=\linewidth]{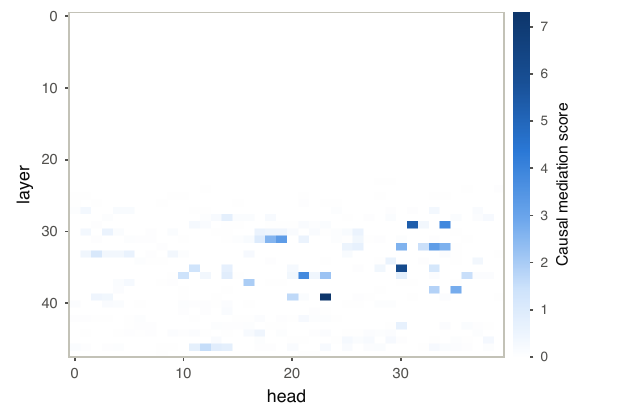}
  \caption{\textit{dual-change belief}, average score}
\end{subfigure}\hfill
\begin{subfigure}[t]{0.49\textwidth}
  \includegraphics[width=\linewidth]{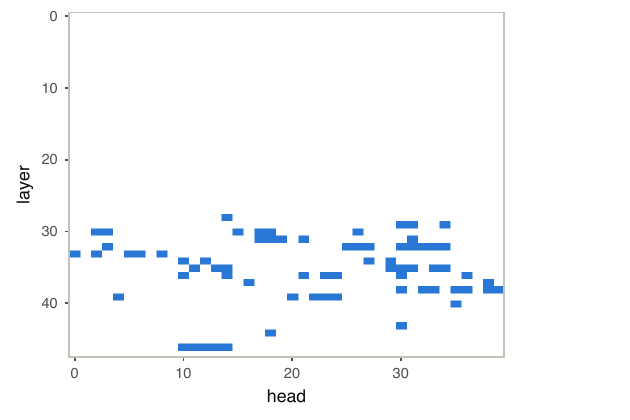}
  \caption{\textit{dual-change belief}, significant heads}
\end{subfigure}
\par\smallskip
\begin{subfigure}[t]{0.49\textwidth}
  \includegraphics[width=\linewidth]{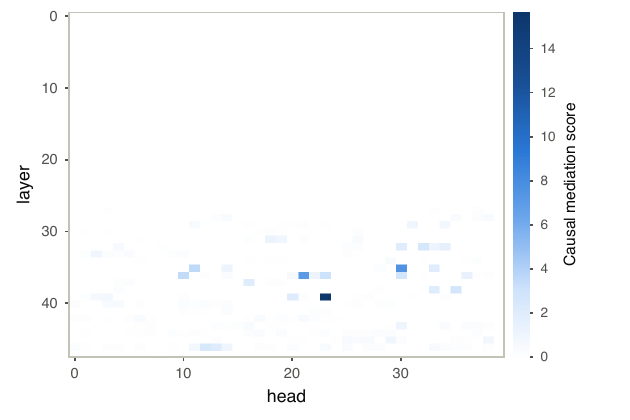}
  \caption{\textit{dual-change photo}, average score}
\end{subfigure}\hfill
\begin{subfigure}[t]{0.49\textwidth}
  \includegraphics[width=\linewidth]{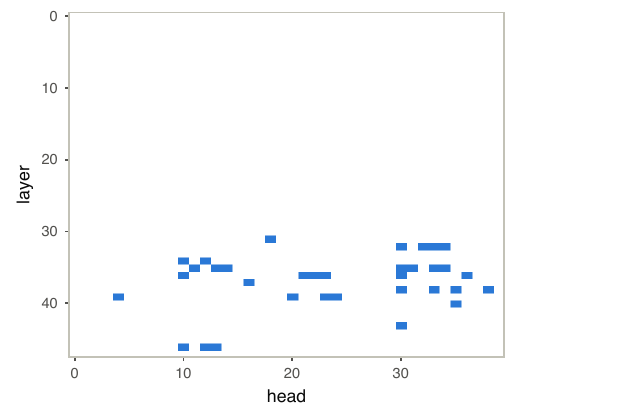}
  \caption{\textit{dual-change photo}, significant heads}
\end{subfigure}
\par\smallskip
\caption{Qwen2.5-14B-Instruct, divergence-change and dual-change conditions. Cross-template averaged CMA scores (left) and FWER-corrected significance masks (right, $\alpha = 0.05$); values in Table~\ref{tab:appendix-heatmaps}.}
\label{fig:appendix-qwen-c14}
\end{figure}

\begin{figure}[tbp]
\centering
\begin{subfigure}[t]{0.49\textwidth}
  \includegraphics[width=\linewidth]{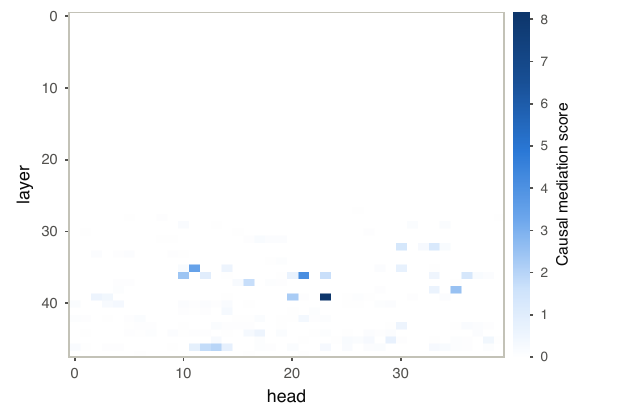}
  \caption{\textit{answer-change belief}, average score}
\end{subfigure}\hfill
\begin{subfigure}[t]{0.49\textwidth}
  \includegraphics[width=\linewidth]{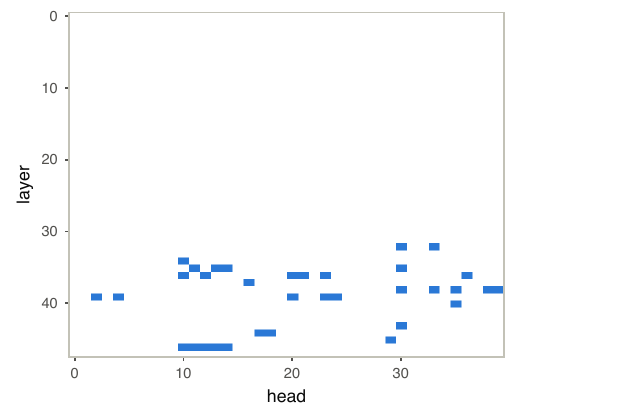}
  \caption{\textit{answer-change belief}, significant heads}
\end{subfigure}
\par\smallskip
\begin{subfigure}[t]{0.49\textwidth}
  \includegraphics[width=\linewidth]{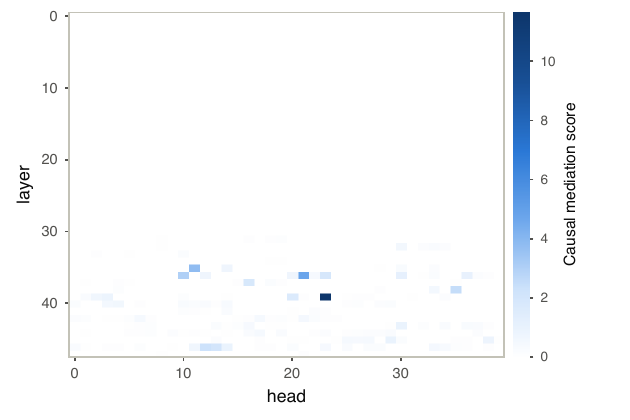}
  \caption{\textit{answer-change photo}, average score}
\end{subfigure}\hfill
\begin{subfigure}[t]{0.49\textwidth}
  \includegraphics[width=\linewidth]{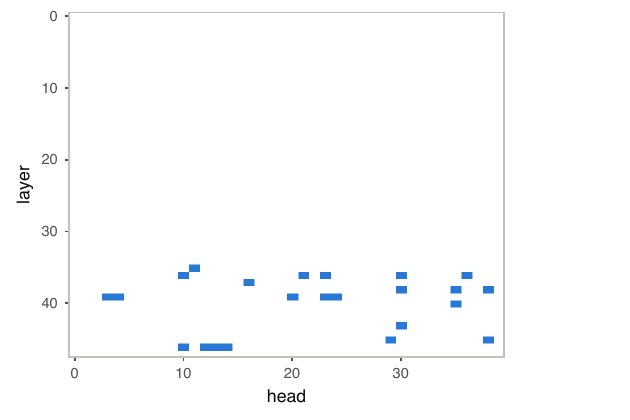}
  \caption{\textit{answer-change photo}, significant heads}
\end{subfigure}
\par\smallskip
\begin{subfigure}[t]{0.49\textwidth}
  \includegraphics[width=\linewidth]{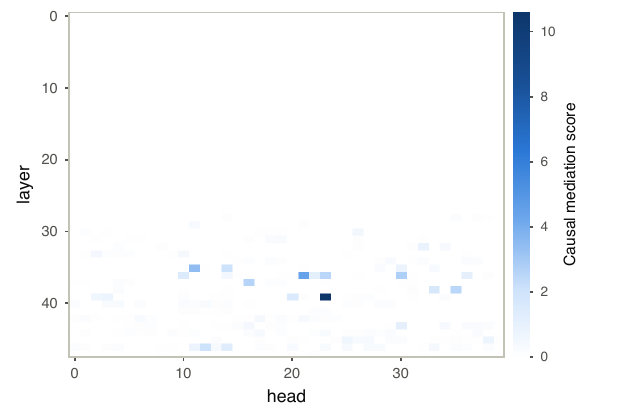}
  \caption{\textit{answer-change state-only}, average score}
\end{subfigure}\hfill
\begin{subfigure}[t]{0.49\textwidth}
  \includegraphics[width=\linewidth]{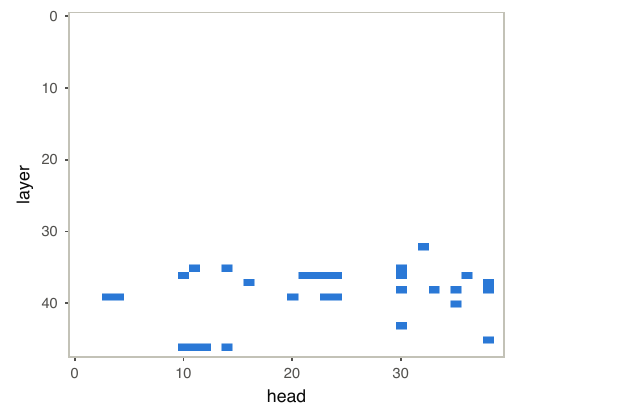}
  \caption{\textit{answer-change state-only}, significant heads}
\end{subfigure}
\par\smallskip
\caption{Qwen2.5-14B-Instruct, answer-change conditions. Layout as in Figure~\ref{fig:appendix-qwen-c14}; values in Table~\ref{tab:appendix-heatmaps}.}
\label{fig:appendix-qwen-c57}
\end{figure}

\subsection{Replication on Gemma-2-27B-it}
\label{sec:appendix-heatmaps-gemma}

We repeated causal mediation analysis on Gemma-2-27B-it (46 layers $\times$ 32 heads, 1472 heads total). \textit{Divergence-change belief}, \textit{divergence-change photo}, \textit{dual-change belief}, and \textit{answer-change belief} each have six cells spanning three templates and two pairing directions. The dual-change analysis uses false/true-belief pairs whose answers also differ, matching the Qwen design; it includes $146$--$150$ valid pairs per cell. The answer-change analysis holds belief status fixed and includes $142$--$150$ valid pairs per cell. Their decomposition gives $r=0.986$ for raw scores and $r=0.985$ for clamped scores. \textit{Divergence-change belief} cells contain $49$--$198$ pairs and \textit{divergence-change photo} cells $150$--$200$. Figure~\ref{fig:appendix-gemma} shows all four conditions: significant divergence-change heads concentrate in a mid-to-upper layer band and overlap across belief and photograph tasks, whereas the \textit{answer-change belief} heads lie later, in the upper layers, and the \textit{dual-change belief} map combines the two bands. The regional localization, the divergence/answer depth ordering and the within-model overlap replicate, without implying correspondence between individual head coordinates across architectures. Residual-stream steering and its downstream attention readout were performed only in Qwen.

\begin{figure}[p]
\centering
\begin{subfigure}[t]{0.49\textwidth}
  \includegraphics[width=\linewidth]{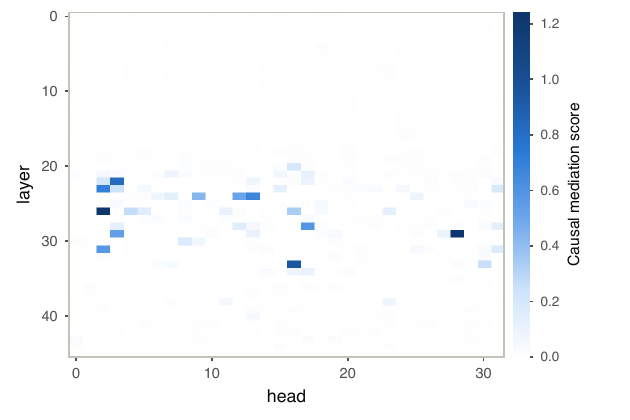}
  \caption{\textit{divergence-change belief}, average score}
\end{subfigure}\hfill
\begin{subfigure}[t]{0.49\textwidth}
  \includegraphics[width=\linewidth]{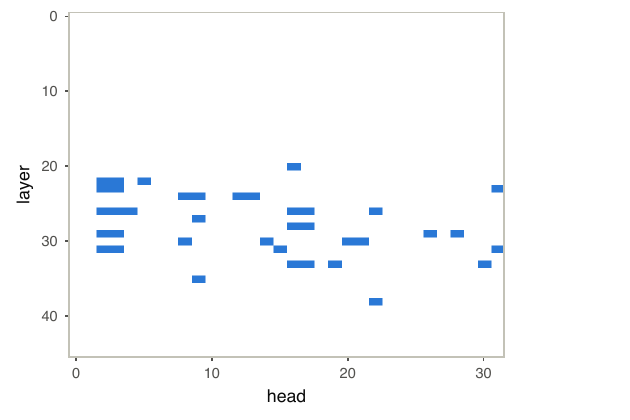}
  \caption{\textit{divergence-change belief}, significant heads}
\end{subfigure}
\par\smallskip
\begin{subfigure}[t]{0.49\textwidth}
  \includegraphics[width=\linewidth]{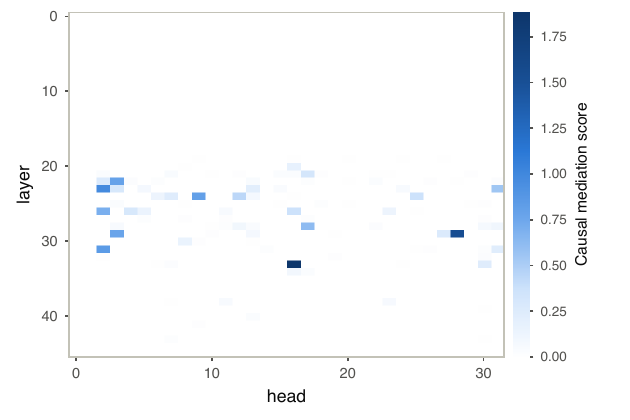}
  \caption{\textit{divergence-change photo}, average score}
\end{subfigure}\hfill
\begin{subfigure}[t]{0.49\textwidth}
  \includegraphics[width=\linewidth]{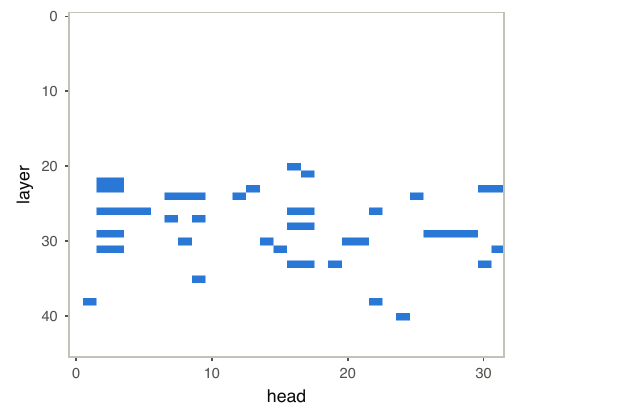}
  \caption{\textit{divergence-change photo}, significant heads}
\end{subfigure}
\par\smallskip
\begin{subfigure}[t]{0.49\textwidth}
  \includegraphics[width=\linewidth]{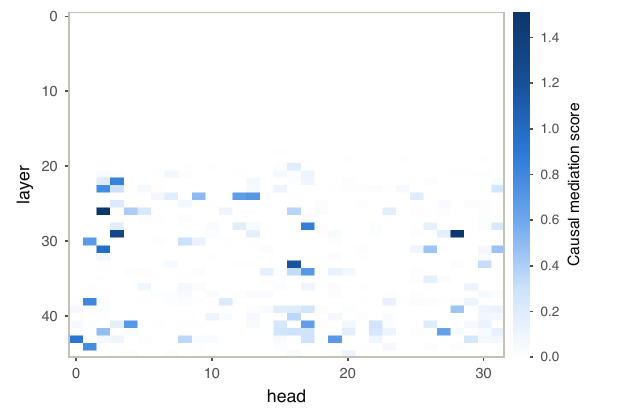}
  \caption{\textit{dual-change belief}, average score}
\end{subfigure}\hfill
\begin{subfigure}[t]{0.49\textwidth}
  \includegraphics[width=\linewidth]{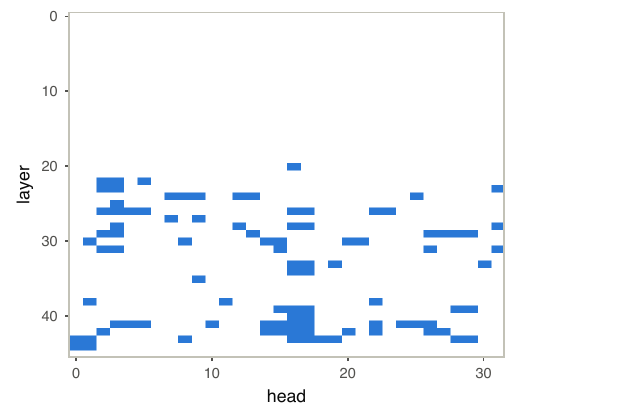}
  \caption{\textit{dual-change belief}, significant heads}
\end{subfigure}
\par\smallskip
\begin{subfigure}[t]{0.49\textwidth}
  \includegraphics[width=\linewidth]{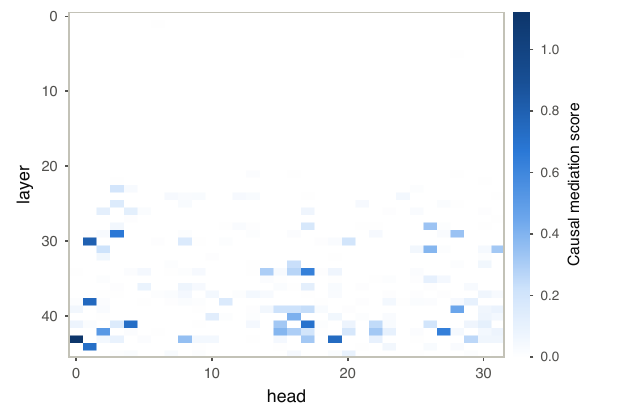}
  \caption{\textit{answer-change belief}, average score}
\end{subfigure}\hfill
\begin{subfigure}[t]{0.49\textwidth}
  \includegraphics[width=\linewidth]{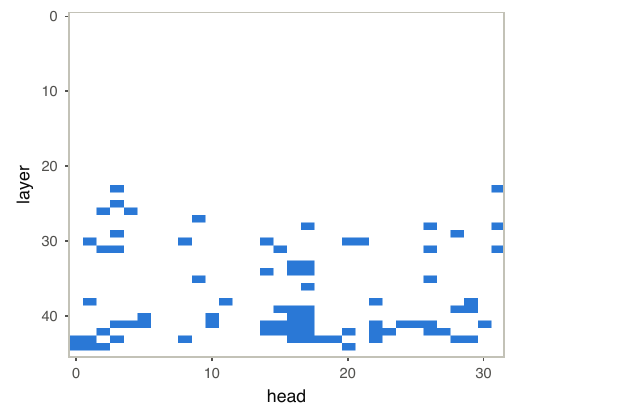}
  \caption{\textit{answer-change belief}, significant heads}
\end{subfigure}
\par\smallskip
\caption{Gemma-2-27B-it, \textit{divergence-change belief}, \textit{divergence-change photo}, \textit{dual-change belief} and \textit{answer-change belief} (46 layers $\times$ 32 heads). Layout matches Figures~\ref{fig:appendix-qwen-c14}--\ref{fig:appendix-qwen-c57}; values in Table~\ref{tab:appendix-heatmaps}.}
\label{fig:appendix-gemma}
\end{figure}
\endgroup

\section{Specificity of the additive decomposition}
\label{sec:appendix-h2}

We test whether adding the separate divergence-change and answer-change fingerprints predicts the dual-change fingerprint better than either component alone. We also ask whether this agreement depends on matching individual heads, or can be explained by their distribution across layers.

\begin{figure}[t]
  \centering
  \includegraphics[width=0.55\textwidth]{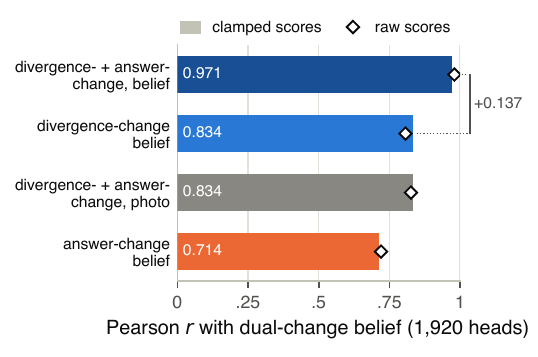}
  \caption{Pearson $r$ with the \textit{dual-change belief} fingerprint over all 1920 heads for the unweighted sum of \textit{divergence-change belief} and \textit{answer-change belief}, each component alone, and the corresponding photo sum. Bars show clamped scores; diamonds show raw scores. The bracket marks the gain of the sum over its better component (95\% head-resampling interval in Section~\ref{sec:modularized}). See Figure~\ref{fig:c3-decomp} for the corresponding scatter plot.}
  \label{fig:c3-specificity}
\end{figure}

\paragraph{Fingerprint construction.} For each story condition, we average each head's causal mediation score over prompt pairs within a cell, then average equally over the six template-by-direction cells. Code fingerprints are averaged over two cells on the indexed-list template: opposite patch directions for divergence-change and dual-change, and copy/alias frames for answer-change. Raw fingerprints keep signed cell means; in clamped fingerprints, negative cell means are set to zero before averaging. Both versions are analyzed in Qwen (1920 heads) and Gemma (1472 heads). The main-text cross-task heatmaps display clamped scores, normalized separately within each panel. The Qwen \textit{divergence-change code} fingerprint uses 148 valid prompt pairs per patch direction. The independently sampled prompts used for the position control are not pooled into this fingerprint. All fingerprint comparisons use every head, without significance filtering.

\paragraph{Per-cell inference.} A max-statistic sign-flip permutation test controls the family-wise error rate across heads within each cell (1920 heads in Qwen, 1472 in Gemma).

\paragraph{Decomposition and uncertainty.} We compare the Pearson correlation of $\mathbf{x}_{\mathrm{dual}}$ with the unweighted sum $\mathbf{x}_{\mathrm{div}}+\mathbf{x}_{\mathrm{ans}}$ against its correlation with either component. The photo sum $\mathbf{x}_{\mathrm{div}}^{\mathrm{photo}}+\mathbf{x}_{\mathrm{ans}}^{\mathrm{photo}}$ combines conditions that vary the same two properties, divergence and answer, in a different task. For the gain over the better component, we resample head indices 10,000 times with replacement, using the same indices for every condition, recomputing both correlations and their maximum in each draw. Intervals are the 2.5th and 97.5th percentiles. These are sensitivity intervals over head identities, which are not independent experimental replicates; they do not estimate uncertainty over new prompts or templates. We also fit $\mathbf{x}_{\mathrm{dual}}=a\mathbf{x}_{\mathrm{div}}+b\mathbf{x}_{\mathrm{ans}}+c\mathbf{1}+\boldsymbol{\epsilon}$ by ordinary least squares, including an intercept, with 10,000 paired head resamples for coefficient intervals.

The fitted coefficients support approximate additivity: the divergence coefficient is near one, while the answer-retrieval coefficient is somewhat below one in both models' belief tasks. In code, both coefficients are close to one (Table~\ref{tab:h2-coefficients}).

\begin{table}[htbp]
  \centering
  \small
  \caption{Coefficients for raw fingerprints. Exact unweighted addition predicts $a=b=1$ and $c=0$. Brackets give 95\% head-resampling intervals from 10,000 resamples.}
  \label{tab:h2-coefficients}
  \begin{tabular}{lccc}
    \toprule
    Model and task & Divergence $a$ & Answer $b$ & Intercept $c$ \\
    \midrule
    Qwen belief & $1.01\ [0.97,1.05]$ & $0.79\ [0.73,0.88]$ & $0.0023$ \\
    Gemma belief & $1.13\ [1.07,1.20]$ & $0.92\ [0.87,0.98]$ & $-0.0063$ \\
    Qwen code & $1.03\ [0.98,1.07]$ & $1.01\ [0.95,1.05]$ & $-0.0029$ \\
    \bottomrule
  \end{tabular}
\end{table}

\paragraph{Residuals.} A residual is the difference between a head's dual-change score and the sum of its divergence-change and answer-change scores. For raw belief fingerprints, larger absolute summed effects tend to have larger absolute residuals in both models ($r=0.69$ in Qwen; $0.58$ in Gemma). Together with the correlation and coefficient results, this supports approximate rather than exact addition of the causal fingerprints.

\paragraph{Permutation controls.} Heads in the same layer may have similar effects. We therefore shuffle head identities within each layer in the summed fingerprint, preserving the layer-level pattern while breaking the correspondence between individual heads. This tests whether the observed agreement exceeds what the layer-level pattern alone would predict. We also shuffle identities across all heads, testing agreement under arbitrary head correspondence. Each control uses 5,000 permutations. In Qwen, the largest within-layer shuffled correlation is $0.451$ for clamped scores, versus the observed $0.971$; the Monte Carlo $p$-value is $1/5001$. The raw result is $r=0.980$, and Gemma gives $r=0.985$ clamped and $0.986$ raw.

\section{Steering methods and additional controls}
\label{sec:steering-methods}

\paragraph{Full-prompt attention profiles.} The separate Qwen attention captures cover 450 prompts per state in each story domain (150 per template), and 24 unpadded, mutated-slot queries per state in code. The retrieval groups contain 16 belief heads, eight photograph heads, and 14 code heads, selected on each domain's answer-change condition and held fixed across the two states. Story heads are significant in at least three of the six template-by-direction cells; code heads are the union of positively significant heads in the two \textit{answer-change code} cells. We average final-token attention over prompts within each template, equally over templates, then over heads weighted by their nonnegative mean causal mediation score in the corresponding retrieval condition. Location spans and the remaining prompt regions form a disjoint partition; absent regions contribute zero. Table~\ref{tab:retrieval-attention-regions} reports descriptive means, with total attention as the denominator. The correct-answer location receives 9.8--26.6 times the mass expected under uniform attention to tokens, where the uniform baseline is the location's fraction of prompt tokens. These captures use different head groups, prompts, and score-weighted averaging from the unweighted 26-head steering readout below.

\begin{table}[ht]
\centering
\small
\begin{tabular}{llrrrrr}
\toprule
Frame & State & Correct & Alternative & Other content & System & End \\
\midrule
Belief & Diverged & 33.9 & 3.8 & 18.0 & 9.9 & 34.5 \\
Belief & Aligned  & 32.8 & 7.0 & 16.7 & 8.8 & 34.6 \\
Photo  & Diverged & 35.1 & 0.6 & 9.6 & 9.6 & 45.0 \\
Photo  & Aligned  & 38.6 & 0.9 & 6.1 & 9.9 & 44.4 \\
Code   & Diverged & 25.0 & 9.4 & 24.1 & 18.2 & 23.3 \\
Code   & Aligned  & 32.2 & 7.6 & 16.7 & 20.6 & 23.0 \\
\bottomrule
\end{tabular}
\caption{Retrieval-head attention as a percentage of total mass. Diverged denotes false belief, an outdated photograph, or a code copy; aligned denotes true belief, an updated photograph, or a code alias. Other content combines all story/code and question tokens outside the two location spans. System includes the prompt prefix; End includes the answer instruction, chat-template tail, and final token. Rows sum to $100\%$ up to rounding.}
\label{tab:retrieval-attention-regions}
\end{table}

\paragraph{Head-output steering.} In Qwen, the 22 heads are the top 32 under the \textit{divergence-change belief} ranking excluding the top 22 under \textit{answer-change belief} (layers 28--36). For each head and task, we estimate a direction from the difference between mean final-token outputs on non-diverged and diverged prompts. We normalize this direction to unit length, then scale it by the length of that head's mean-difference vector in the target task before adding it to the head output. The sign is chosen per prompt so that $\alpha=+1$ corresponds to an edit toward the opposite status and $\alpha=-1$ to an edit toward its original status. For cross-task steering, the direction is estimated from source-task examples and its magnitude from target-task activations. For within-task steering, direction fitting and evaluation use the same prompts. Evaluation includes 72 belief prompts, 72 photograph prompts, and 96 code prompts. Controls are matched-norm random directions and the estimated directions shuffled across the same heads. The readout is the intervention-induced change in the alternative-minus-correct logit margin, minus that produced by the random control. The effects of all evaluated cross-task directions at $\alpha=+1$ differ from those of their matched random control (paired Wilcoxon signed-rank tests, $p<10^{-12}$).

Writing $d_h^S$ and $d_h^T$ for head $h$'s non-diverged-minus-diverged mean differences in source and target tasks, respectively, the head edit is
\[
z'_h(p)=z_h(p)+\alpha s_p\|d_h^T\|\frac{d_h^S}{\|d_h^S\|},
\]
where $s_p=+1$ for a diverged prompt and $s_p=-1$ otherwise. This expresses the figure's sign convention: positive $\alpha$ corresponds to an edit toward the opposite status. The norm-matched random control replaces the final unit vector with a random unit vector independently for each prompt and head.

\begin{figure}[t]
  \centering
  \includegraphics[width=\textwidth]{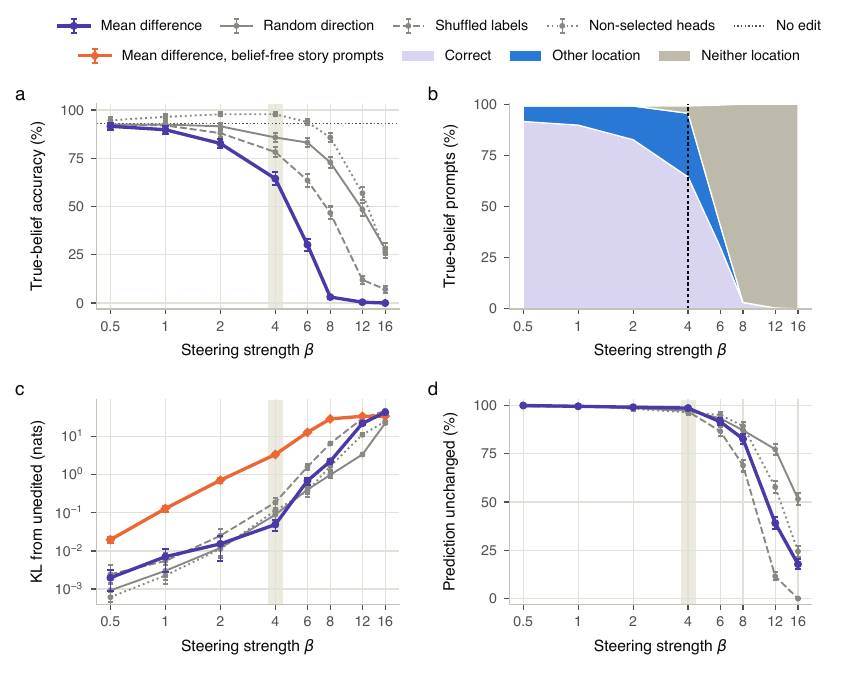}
  \caption{Choosing the residual steering strength $\beta$: the mean-difference edit versus norm-matched control directions (grey). (a) Accuracy on true-belief prompts when each direction is added. (b) The answer predicted on true-belief prompts under the mean-difference edit: correct, the other story location (the false-belief answer), or neither location (the remaining ${\le}1.3\%$ begin both locations); dashed line, $\beta=4$. (c) KL divergence of the edited from the unedited next-token distribution under addition, on generic belief-free prompts and, for the mean-difference edit, on belief-free story prompts. (d) Generic prompts whose most probable next token is unchanged. Error bars, $\pm1$ standard error over prompts.}
  \label{fig:coefficient-sweep}
\end{figure}

\paragraph{Residual-stream steering.} This separate experiment uses Qwen belief prompts, with a fixed split within each of the three templates: half for fitting, half for evaluation (225 held-out pairs in total). Let $d=\mu_{\mathrm{FB}}-\mu_{\mathrm{TB}}$ be the difference of fit-half mean residual states after layer 41 at the final token. We denote residual-steering strength by $\beta$, distinguishing it from the head-output coefficient $\alpha$. Single-direction injection is $x'=x+\beta d$ and deletion is $x'=x-\beta d$. At $\beta=1$, the edit adds or subtracts one difference of the two condition means; it does not replace the prompt's state with a condition mean. We evaluate $\beta\in\{0.5,1,2,4,6,8,12,16\}$ and an unedited baseline on 225 held-out prompts per condition. Figure~\ref{fig:coefficient-sweep} reports the coefficient comparisons; the main-text attention analysis uses $\beta=4$. This adapts the mean-difference additive intervention of \citet{rimskySteeringLlama2024}; unlike their generation-wide intervention, we edit only the final prompt token and measure both the next answer token and downstream attention. The edit occurs at the input to layer 42, without directly editing the 26 measured retrieval heads in layers 42--46.

\paragraph{Steering-strength selection.} We selected $\beta=4$ from the parameter sweep to balance intervention effects with output stability. At this strength, addition on true-belief prompts produces the highest alternative-location response rate among the tested coefficients ($31.1\%$), with only $3.6\%$ of responses outside the two candidate locations. At $\beta=6$, these rates are $10.7\%$ and $58.7\%$, respectively. Subtraction on false-belief prompts responds differently across coefficients: at $\beta=4$, $5.3\%$ select the alternative and $0.4\%$ select neither candidate, whereas its alternative-response rate peaks at $\beta=8$, with $32.4\%$ outside the candidate pair. At the selected strength, query restoration significantly reverses the steering-induced answer-margin shift beyond matched controls, as reported below.

\paragraph{Exploratory output-stability control.} We tested output stability on a random sample of 225 custom belief-free prompts spanning factual questions, arithmetic, language, and code tasks. At $\beta=4$, addition and subtraction preserve the unedited model's highest-probability next token on $98.7\%$ and $91.1\%$ of prompts, respectively (Figure~\ref{fig:coefficient-sweep}).

\paragraph{Residual controls.} We draw a Gaussian random unit vector, recompute the mean difference after permuting false/true labels within each template's fit half, and sum fit-half $\Delta z_h W_O^{(h)}$ vectors from 30 non-selected heads matched to the selected heads' layers. Each direction is normalized and added at the same layer and token with norm $\beta\|d\|$. These controls match the single-direction operator and norm.

\paragraph{Behavioral readout.} Accuracy is the fraction of prompts whose highest-logit next token matches the beginning of the correct answer and not the alternative; there are no first-token collisions in this stimulus set. Table~\ref{tab:residual-controls} compares the target edits and controls.
Under single-direction steering at $\beta=4$, deletion on true belief gives $94.2\%$ accuracy, and injection on false belief gives $83.1\%$.
On 225 basic location prompts without a belief variable, mean full-vocabulary probability of the correct answer token falls from $0.900$ clean to $0.634$ under single-direction injection at $\beta=4$. These probabilities range from zero to one and are not accuracy or probabilities renormalized over the two candidate answers.

\begin{table}[ht]
\centering
\small
\begin{tabular}{lrr}
\toprule
Intervention & False belief: deletion & True belief: injection \\
\midrule
Clean & 96.9 & 92.9 \\
Single-direction mean difference & 94.2 & 64.4 \\
Random direction & 94.7 & 85.8 \\
Shuffled-label direction & 92.0 & 78.2 \\
Non-selected-head direction & 93.3 & 97.8 \\
\bottomrule
\end{tabular}
\caption{First-token accuracy (\%) on 225 held-out prompts per belief condition, with $\beta=4$ for every edit. Random, shuffled-label, and non-selected-head controls match the single-direction edit's norm and additive operator.
}
\label{tab:residual-controls}
\end{table}

\paragraph{Attention readout and baseline preference.} Before steering, the retrieval heads attend more strongly on average to the correct location than to the alternative. We measure attention from the final prompt token to tokens matching each location name, summing over matched tokens and averaging across the 26 retrieval heads within each prompt. These values are fractions of total attention across the prompt, not renormalized over the two locations. On false-belief prompts, the correct and alternative locations receive an average of $11.50\%$ and $3.54\%$ of attention, respectively; on true-belief prompts, they receive $12.00\%$ and $6.79\%$. Correct and alternative locations remain defined by the original story throughout the steering analysis.

The location-matching procedure also includes text outside the story or partial name matches in four of the 225 held-out pairs. Excluding those pairs leaves 221 pairs and essentially unchanged correct-minus-alternative differences: $7.94$ percentage points for false belief and $5.23$ for true belief. The plotted bars include all pairs. Because the stored attention measurements sum over matched tokens, this sensitivity check excludes affected pairs rather than reconstructing the matches. The separate full-prompt profiles above provide the comparison with other prompt regions.

\paragraph{Retrieval-head query restoration and transplantation.}
\label{sec:query-restoration}
We tested whether steering changes the answer partly through the retrieval heads' queries. In the restoration intervention, we kept residual steering active but restored these queries to their unedited values, testing whether this reverses part of the answer shift. In the transplantation intervention, we inserted the queries from a steered run into an otherwise unedited run, testing whether these query changes can reproduce part of the shift.

We used Qwen2.5-14B-Instruct with the same fitted direction, 26 retrieval heads in layers 42--46, and 225 held-out prompts per belief condition (75 per template) as the attention experiment. The residual edit acts only at the final prompt token, at the input to layer 42, with $\beta=4$: subtracting the false-minus-true direction on false-belief prompts and adding it on true-belief prompts. For each prompt, we recorded the heads' final-token query vectors in unedited and steered runs. Each replacement occurred at the head's own layer, and subsequent computation proceeded normally. Control interventions restored unedited queries in three randomly sampled sets of 26 heads outside the selected retrieval set. Each control set matched the retrieval set's head count within every layer and was fixed across prompts. We averaged the three control recoveries within each prompt before comparing them with retrieval-head recovery.

The readout is the model's final next-token margin $m=\operatorname{logit}(y_{\mathrm{alt}})-\operatorname{logit}(y_{\mathrm{correct}})$, after the last transformer layer and output projection. Let $m_0$, $m_S$, $m_R$, and $m_T$ denote the margins in the unedited, steered, retrieval-query restoration, and retrieval-query transplantation runs, and let $m_{R,j}$ denote control-restoration run $j$. We compute, per prompt,
\begin{align*}
\text{steering effect} &= m_S-m_0, &
\text{retrieval-query recovery} &= m_S-m_R,\\
\text{control recovery} &= \frac{1}{3}\sum_{j=1}^{3}(m_S-m_{R,j}), &
\text{transplant effect} &= m_T-m_0.
\end{align*}
Recovery beyond controls is the retrieval-query recovery minus the control recovery, paired within prompt. Positive recovery means that restoration reverses part of the steering-induced margin shift.

\begin{table}[htbp]
\centering
\small
\begin{tabular}{lrr}
\toprule
Quantity & False belief (subtract) & True belief (add) \\
\midrule
Steering effect & $13.49\pm0.43$ & $5.37\pm0.40$ \\
Retrieval-query recovery & $2.80\pm0.14$ & $3.11\pm0.22$ \\
Control-query recovery & $0.71\pm0.05$ & $0.36\pm0.07$ \\
Recovery beyond controls & $2.09\pm0.15$ & $2.76\pm0.24$ \\
Retrieval-query transplant effect & $4.07\pm0.22$ & $1.75\pm0.15$ \\
\bottomrule
\end{tabular}
\caption{Query restoration and transplantation: mean answer-margin effects in logits $\pm$ one standard error over 225 held-out prompts per condition. Control recovery is averaged over three layer-matched random head sets within each prompt; its uncertainty is over prompts, conditional on these sampled sets.}
\label{tab:query-restoration}
\end{table}

Retrieval-query restoration recovered 20.8\% and 58.0\% of the respective mean steering effects, compared with 5.2\% and 6.7\% for controls (Table~\ref{tab:query-restoration}). Recovery beyond controls was therefore equivalent to 15.5\% and 51.3\% of the mean steering effects; transplantation reproduced 30.1\% and 32.6\%. Two-sided paired Wilcoxon signed-rank tests of retrieval-minus-control recovery against a distribution symmetric about zero gave nominal $p=3.2\times10^{-29}$ and $p=1.7\times10^{-21}$, respectively.

The percentages above are ratios of condition means, not averages of per-prompt percentages. Restoration and transplantation are separate interventions whose effects need not add to the full steering effect.

These interventions support a partial causal contribution through the retrieval heads' queries. Other routes and interactions can contribute to the remaining answer-margin shift. The interventions replace complete query vectors, so they do not isolate the contribution of attention to the two location spans alone.

\section{Representational similarity and position controls}
\label{sec:per-cond-rdm}

Causal mediation measures a head's effect on the answer. To characterize the information in its outputs, we used representational similarity analysis following \citet{yangEmergentSymbolicMechanisms2025}. We tested whether heads selected on divergence-change conditions encode divergence status and whether heads selected on answer-change conditions encode answer identity.

\paragraph{Rationale.} In our story templates the false-belief answer is always the first location mentioned and the true-belief answer the second, so a model RDM for belief veracity and a model RDM for the answer's position in the text correlate at $r = 0.996$--$0.999$ within each template. This implies that fit to the veracity model cannot be told apart from a fit to answer position. We therefore use the Python analog from the code conditions, in which position can be varied independently of veracity: after \texttt{a[i]} is mutated, \texttt{b = a.copy()} leaves \texttt{b[i]} at its original value (diverged, the false-belief analog), whereas \texttt{b = a} exposes the update (not diverged, the true-belief analog).

\paragraph{Head selection.} We test four Qwen2.5-14B-Instruct head sets. The code divergence set contains 12 heads with positive, family-wise-error-corrected significant scores in either direction of the \textit{divergence-change code} condition. The code retrieval set contains 14 heads selected by the same rule on the \textit{answer-change code} condition. The story divergence and story retrieval sets contain 32 and 22 heads, respectively, selected on the \textit{divergence-change belief} and \textit{answer-change belief} fingerprints. Because the story sets were selected on story prompts, evaluating them on code tests whether their representational structure extends across domains.

\paragraph{Comparing representational structure.} We ask whether prompts that share a feature, such as divergence status or answer identity, produce similar head outputs. For each stimulus, we ran a single forward pass with the chat template and code system prompt and captured each head's output $z$ at the final prompt token, before the attention output projection. We built a representational dissimilarity matrix (RDM) for each head, with one row and column per stimulus and cosine-distance entries $(i,j)=1-\cos(z_i,z_j)$. A head set's RDM is the unweighted mean of its heads' RDMs.

We compare these observed RDMs with model RDMs describing differences in the tested features. The comparison is the Spearman correlation between their upper triangles, also called the Mantel correlation (range $-1$ to $1$; zero indicates no rank correlation). Significance is assessed with $1000$ stimulus-label permutations ($p$-floor $0.001$). Noise ceilings estimate how reliably the measured structure can be reproduced using replicate stimuli within each design cell. We compute them as the square root of Spearman--Brown-corrected replicate reliability. Because they are computed at the cell level, they are optimistic.

\paragraph{Experiment 1: divergence versus position.} In the story prompts, divergence status and answer position are closely linked. Experiment 1 uses code prompts to vary the queried list slot and surrounding text separately from copy/alias status. We ask whether the heads' output similarities follow divergence status or these positional features.

\begin{figure}[tbp]
  \centering
  \includegraphics[width=\textwidth]{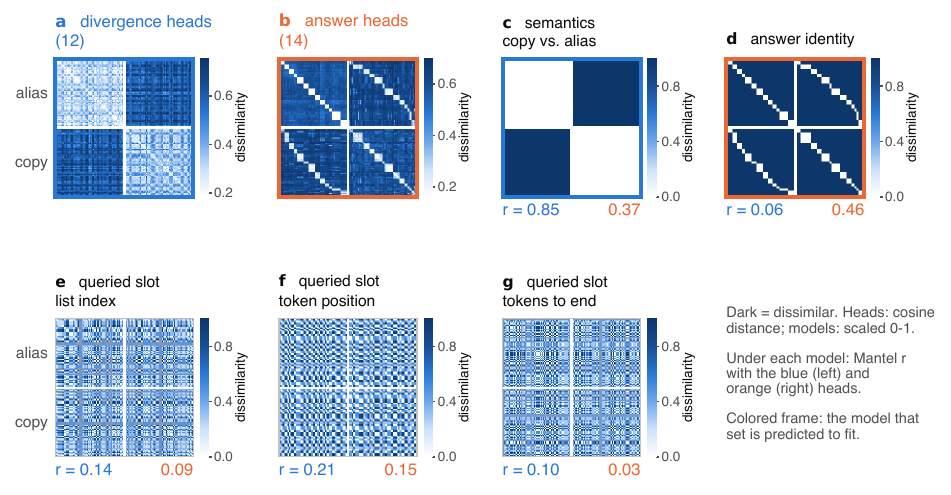}
  \caption{Experiment 1 RDMs (Qwen2.5-14B-Instruct, 144 code prompts). \textbf{(a, b)} Observed RDMs: cosine distance averaged over the code divergence set (blue) or code retrieval set (orange). \textbf{(c--g)} Model RDMs. Stimuli are ordered alias, then copy, each by answer value and queried slot position. Each model RDM is divided by its maximum, so all model panels share a 0--1 scale (position models are differences in tokens or list indices before scaling; Mantel correlations are rank-based and unaffected). Under each model: Mantel correlation with (a) in blue and (b) in orange (Table~\ref{tab:rdm-exp1}). Colored frames mark the model each set is predicted to fit.}
  \label{fig:rdm-exp1}
\end{figure}

The 144 single-object prompts cross copy/alias status, queried list slot (indices 1--4), prefix padding, and suffix padding, with two replicates per combination. Padding consists of an unrelated list containing 0, 16, or 32 filler elements before the code or after the mutation; these independently shift absolute position and distance to the end. The positional models describe the queried slot's index in the list, its absolute token position in the initial \texttt{a = [...]} line, and its distance from the end of the prompt. These slot-position measures differ from the position of the token supplying the correct answer: copy answers come from the initial list, whereas alias answers come from the mutation line. On the tokenized stimuli, the copy/alias-status model correlates with the queried-slot position models at $r\approx-0.01$ and with answer-token position at $r=0.086$. We compare the head-output RDMs with the copy/alias-status, slot-position, and answer-identity models (Table~\ref{tab:rdm-exp1}, Figure~\ref{fig:rdm-exp1}).

\begin{table}[H]
  \caption{Experiment 1 (position grid, 144 stimuli): Mantel correlations between each head set's RDM and each model RDM. The largest null value over $1000$ permutations is $\approx 0.04$--$0.07$; every cell reaches $p = 0.001$ except the code retrieval set $\times$ distance to end ($p = 0.034$). Noise ceilings: code divergence $0.99$, story divergence $0.97$, story retrieval $0.85$, code retrieval $0.63$. Bold marks the predicted model for each head set (copy/alias status for divergence sets, answer identity for retrieval sets), not a significance threshold.}
  \label{tab:rdm-exp1}
  \centering
  \small
  \begin{tabular}{lccccc}
    \toprule
    Head set (size) & copy/alias & \shortstack{list\\index} & \shortstack{token\\position} & \shortstack{tokens\\to end} & \shortstack{answer\\identity} \\
    \midrule
    code divergence (12) & \textbf{0.851} & 0.142 & 0.209 & 0.095 & 0.059 \\
    story divergence (32) & \textbf{0.775} & 0.177 & 0.299 & 0.105 & 0.149 \\
    code retrieval (14) & 0.373 & 0.090 & 0.155 & 0.030 & \textbf{0.461} \\
    story retrieval (22) & 0.613 & 0.122 & 0.233 & 0.064 & \textbf{0.448} \\
    \bottomrule
  \end{tabular}
\end{table}

\begin{figure}[tbp]
  \centering
  \includegraphics[width=\textwidth]{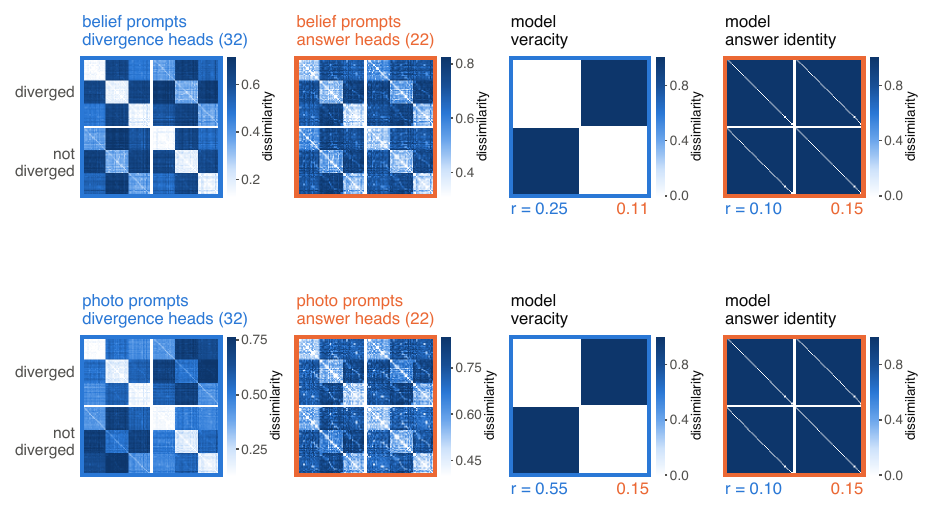}
  \caption{Representational dissimilarity matrices for belief (top) and photograph (bottom) prompts in Qwen2.5-14B-Instruct: 144 prompts per task, 48 per template. Each task's top 32 divergence heads and top 22 retrieval heads are compared with veracity and answer-identity model matrices. Stimuli are ordered by veracity, template, then answer. Printed values are Spearman correlations (range $-1$ to $1$); all reach $p=0.001$ under 1000 stimulus-label permutations.}
  \label{fig:rdm-story}
\end{figure}

The code divergence and story divergence sets have their strongest correlations with copy/alias status among the tested models. The code retrieval set fits answer identity most strongly, but the story-selected retrieval set has a higher raw correlation with copy/alias status than with answer identity ($0.613$ versus $0.448$). These raw correlations do not establish stronger selectivity for copy/alias status: the binary model matrices have different proportions of tied entries, so their attainable correlations with distinct observed distance ranks differ. Figure~\ref{fig:rdm-story} shows the descriptive belief and photograph counterparts.

Two questions remain. First, copy and alias prompts differ in syntax, so their separation could partly reflect how the programs are written. Second, this result does not tell us whether the heads distinguish divergence of the list as a whole or divergence of the particular element being queried. Experiment 2 varies the expression of copying and separates these two forms of divergence.

\paragraph{Experiment 2: object and queried-value divergence.} A copied list can differ from the updated original even when the particular element being queried is unchanged. For example, if only the first element changes, the copy and original differ as lists but still agree at the second element. Experiment 2 asks whether the heads distinguish divergence of the list as a whole (\emph{object divergence}) or divergence of the queried element (\emph{slot divergence}).

In each of the 144 prompts, two lists are derived from \texttt{a}, one copy and one alias. We vary copy syntax (\texttt{a.copy()}, \texttt{a[:]}, \texttt{list(a)}) to test whether the representational pattern follows a particular expression. We swap which variable is the copy---\texttt{b} in half the stimuli and \texttt{c} in the other half---to separate object role from variable name. We query either list at mutated and untouched slots to distinguish whole-list divergence from queried-element divergence. Prefix and suffix padding lists contain 0, 16, or 32 filler elements.

We compare the head-output RDMs with models of object divergence, slot divergence, the queried name (\texttt{b} vs.\ \texttt{c}), copy and alias syntax, answer identity, and the queried slot's absolute token position in the initial list. Object divergence identifies whether the queried list is the copy; slot divergence identifies whether its queried value differs from \texttt{a}'s. Slot divergence implies object divergence, so these model RDMs are correlated ($r\approx0.25$). We therefore measure each model's association with the head-output RDM after accounting for the other model. These partial Spearman correlations are computed by rank-residualizing out the other model and are denoted obj$\mid$slot and slot$\mid$obj.

\begin{table}[H]
  \caption{Experiment 2 (object and queried-value divergence, 144 stimuli): Mantel correlations, as in Table~\ref{tab:rdm-exp1}. The largest null value over $1000$ permutations is $\approx 0.04$--$0.06$. Noise ceilings: code divergence $0.59$, story divergence $0.48$. Bold marks the predicted models for each head set (object divergence and obj$\mid$slot for divergence sets, answer identity for retrieval sets).}
  \label{tab:rdm-exp2}
  \centering
  \footnotesize
  \setlength{\tabcolsep}{3pt}
  \resizebox{\textwidth}{!}{%
  \begin{tabular}{lcccccccc}
    \toprule
    Head set (size) & \shortstack{object\\divergence} & \shortstack{slot\\divergence} & slot$\mid$obj & obj$\mid$slot & \shortstack{queried\\name} & \shortstack{syntax\\(copy / alias)} & \shortstack{answer\\identity} & \shortstack{token\\position} \\
    \midrule
    code divergence (12) & \textbf{0.294} & 0.089 & $+$0.015 & \textbf{$+$0.282} & 0.050 & 0.006 / 0.042 & 0.097 & 0.405 \\
    story divergence (32) & \textbf{0.231} & 0.103 & $+$0.047 & \textbf{$+$0.213} & 0.041 & 0.003 / 0.041 & 0.176 & 0.496 \\
    code retrieval (14) & 0.053 & $-$0.046 & $-$0.061 & $+$0.067 & $-$0.010 & 0.000 / 0.000 & \textbf{0.453} & 0.182 \\
    story retrieval (22) & 0.105 & 0.011 & $-$0.016 & $+$0.105 & 0.010 & 0.000 / 0.009 & \textbf{0.424} & 0.343 \\
    \bottomrule
  \end{tabular}}
\end{table}

\begin{figure}[tbp]
  \centering
  \includegraphics[width=\textwidth]{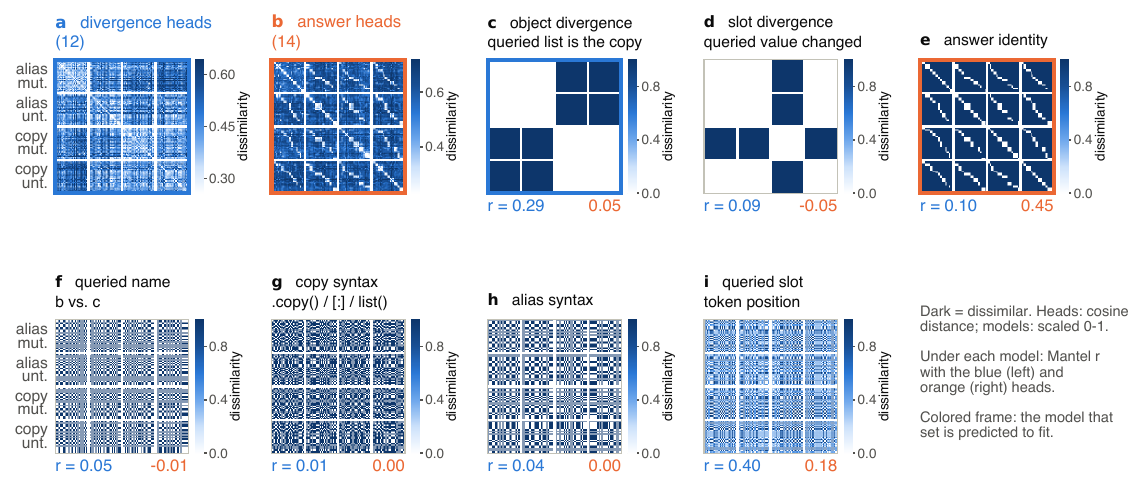}
  \caption{Experiment 2 RDMs. Stimuli are ordered by the queried list (alias, copy), then by whether the queried slot was mutated (mut.) or untouched (unt.), then by answer value. \textbf{(a, b)} Observed RDMs. \textbf{(c, d)} The two divergence models: whether the queried list is the copy (object divergence) and whether the queried value itself changed (slot divergence). \textbf{(e)} Answer identity. \textbf{(f--h)} Surface models. \textbf{(i)} The queried slot's token position in the initial list. The divergence set fits object divergence (c, $r = 0.29$) more strongly than slot divergence or the tested name and syntax models, while the retrieval set fits answer identity (e, $r = 0.45$). Position (i) is fit by both sets and is discussed in the text.}
  \label{fig:rdm-exp2}
\end{figure}

Both divergence sets retain an association with object divergence after accounting for slot divergence (both $p=0.001$). The reverse association, with slot divergence after accounting for object divergence, is not significant ($p=0.65$ for the code set and $p=0.12$ for the story set). Thus, the outputs distinguish whether the list as a whole diverges beyond whether the queried value differs. These partial correlations control for the other divergence model, not for token position. Table~\ref{tab:rdm-exp2} reports the correlations; the two obj$\mid$slot values are $3.7$--$5.1\times$ the largest of their respective $1000$ null draws.

The queried-name and syntax models have weaker correlations with the divergence sets. Copy syntax is not significant in any set ($p\geq0.29$), while the small alias-syntax associations in the divergence sets are significant ($p=0.001$). Both retrieval sets fit answer identity most strongly. They also retain small associations with object divergence after accounting for slot divergence (both $p=0.001$), but these correlations are about two to four times larger in the corresponding divergence sets.

\paragraph{Causal control for absolute answer position.}
\label{sec:position-matched-control}
Experiment 2 finds an association with whole-list divergence after accounting for whether the queried element differs. However, the queried slot's position in the initial list also correlates with all four Qwen head sets ($0.18$--$0.50$) and gives the largest listed correlation for the two divergence sets. Representational similarity alone does not tell us whether the heads' effects on the answer depend on differences in answer position. We therefore test their causal effects while holding the position of the token supplying the correct answer fixed.

We constructed Qwen copy/alias prompt pairs with the same correct answer, the same absolute position of the token supplying that answer, and the same total prompt length. The answer is supplied by the initial list in the copy prompt and by the mutation in the alias prompt. One-element lists eliminate variation in the queried list index, while prefix and suffix comments align the answer-token position and prompt length, also fixing the answer token's distance from the final token. An unmatched comparison uses the same filler budget without aligning the answer position. We then patch individual final-token head outputs to test whether transferring divergence-head outputs still changes the model's answer preference. We test both directions using the same 32 belief-selected and 12 code-selected divergence heads as above, with head selection preceding evaluation on 40 independently generated prompt pairs per direction and matching condition.

The readout is the per-head change in the causal-minus-original-answer logit margin, averaged over the selected heads and prompts; it is signed and unbounded, with zero denoting no effect. Table~\ref{tab:position-matched-control} gives 95\% percentile intervals from 10,000 prompt resamples, testing whether the set-mean effect differs from zero. Matched effects remain positive in both directions for both sets. Matched/unmatched effect ratios have bootstrap medians of $0.89$--$0.99$; their intervals include one but do not establish equivalence.

\begin{table}[htbp]
\centering
\small
\begin{tabular}{llll}
\toprule
Head set & Patch direction & \shortstack{Matched effect\\{[95\% interval]}} & \shortstack{Matched/unmatched\\{[95\% interval]}} \\
\midrule
Belief divergence (32) & copy $\to$ alias & 0.395 [0.301, 0.492] & 0.99 [0.71, 1.35] \\
Code divergence (12) & copy $\to$ alias & 0.843 [0.662, 1.032] & 0.97 [0.72, 1.28] \\
Belief divergence (32) & alias $\to$ copy & 1.259 [1.150, 1.367] & 0.89 [0.79, 1.01] \\
Code divergence (12) & alias $\to$ copy & 2.876 [2.666, 3.086] & 0.92 [0.83, 1.03] \\
\bottomrule
\end{tabular}
\caption{One-element-list position control in Qwen2.5-14B-Instruct. Matched effects are mean causal-mediation scores in logits; ratios are dimensionless bootstrap medians. Intervals resample prompts with the head sets fixed; ratio intervals resample matched and unmatched prompts independently. Each cell contains 40 prompt pairs.}
\label{tab:position-matched-control}
\end{table}

\begin{samepage}
The heads' outputs distinguish whole-list divergence after accounting for queried-value divergence. Their intervention effects persist when absolute answer-token position is matched, so those effects do not require a difference in that position between the paired prompts. Combined with the steering and query-restoration results, these findings support an indirection mechanism in which divergence-sensitive information guides downstream answer retrieval.
\end{samepage}

\paragraph{Gemma replication.} We also evaluated Gemma-2-27B-it story head sets on the code prompts in the two representational experiments. In Experiment 1, the \textit{divergence-change belief}, \textit{divergence-change photo} and \textit{dual-change belief} sets fit copy/alias status at $0.61$--$0.69$, and the \textit{answer-change belief} set fits answer identity at $0.44$. In Experiment 2, the three divergence-change and dual-change sets show the same asymmetry as in Qwen: obj$\mid$slot $= +0.20$, $+0.17$ and $+0.19$, respectively ($p = 0.001$), and slot$\mid$obj between $-0.06$ and $-0.05$.

\paragraph{Caveats.} Experiment 2 varies copy syntax and balances object roles, while the causal control uses independently sampled prompts with matched absolute answer-token positions. These complementary controls strengthen the divergence interpretation, but do not establish a representation free of positional or syntactic information.

\end{document}